\pdfoutput=1
\documentclass{article}

\usepackage[final, nonatbib]{neurips_2021}
\usepackage[dvipsnames]{xcolor}
\usepackage{array,multirow,graphicx}

\newcommand{\method}{Subgoal Search}
\newcommand{\abbrv}{kSubS}

\usepackage[utf8]{inputenc} 
\usepackage[T1]{fontenc}    
\usepackage{hyperref}       
\usepackage{url}            
\usepackage{booktabs}       
\usepackage{amsfonts}       
\usepackage{nicefrac}       
\usepackage{microtype}      
\usepackage{xcolor}         
\usepackage{subfig}
\usepackage{ulem}

\usepackage{algorithmicx}
\usepackage{algorithm}
\usepackage[noend]{algpseudocode}

\usepackage{wrapfig}
\usepackage{graphicx}

\usepackage{amsmath,amsfonts,amsthm,amssymb}

\title{Subgoal Search For Complex Reasoning Tasks}

\author{%
  Konrad Czechowski\thanks{equal contribution} \\
  University of Warsaw \\
  \texttt{k.czechowski@mimuw.edu.pl} \\
  \And
  Tomasz Odrzygóźdź\footnotemark[1] \\
  University of Warsaw \\
  \texttt{tomaszo@impan.pl} \\
  \AND
    Marek Zbysiński \\
  University of Warsaw \\
  \texttt{m.zbysinski@} \\ 
  \texttt{students.mimuw.edu.pl} \\
  \And
    Michał Zawalski \\
  University of Warsaw \\
  \texttt{m.zawalski@uw.edu.pl} \\
  \And
  Krzysztof Olejnik \\
  University of Warsaw \\
  \texttt{k.olejnik3@} \\
  \texttt{student.uw.edu.pl} \\
  \And
  Yuhuai Wu \\
  University of Toronto, \\
  Vector Institute \\
  \texttt{~~~ywu@cs.toronto.edu~~~} \\
  \And
  Łukasz Kuciński \\
  Polish Academy of Sciences \\
  \texttt{lkucinski@impan.pl} \\
  \And
  Piotr Miłoś \\
  Polish Academy of Sciences,\\
  University of Oxford,\\
  deepsense.ai\\
  \texttt{pmilos@impan.pl}
}

\begin{document}

\maketitle

\begin{abstract}
    Humans excel in solving complex reasoning tasks through a mental process of moving from one idea to a related one. Inspired by this, we propose \method{} (\abbrv) method. Its key component is a learned subgoal generator that produces a diversity of subgoals that are both achievable and closer to the solution. Using subgoals reduces the search space and induces a high-level search graph suitable for efficient planning. In this paper, we implement \abbrv{} using a transformer-based subgoal module coupled with the classical best-first search framework. We show that a simple approach of generating $k$-th step ahead subgoals is surprisingly efficient on three challenging domains: two popular puzzle games, Sokoban and the Rubik's Cube, and an inequality proving benchmark INT. \abbrv{} achieves strong results including state-of-the-art on INT within a modest computational budget.
\end{abstract}

\section{Introduction}\label{sec:introduction}

Reasoning is often regarded as a defining property of advanced intelligence \cite{russell2002artificial, hassabis2017neuroscience}. When confronted with a complicated task, humans' thinking process often moves from one idea to a related idea, and the progress is made through milestones, or \textit{subgoals}, rather than through atomic actions that are necessary to transition between subgoals \cite{gowers2000importance}. During this process, thinking about one subgoal can lead to a possibly diverse set of subsequent subgoals that are conceptually reachable and make a promising step towards the problem's solution. This intuitive introspection is backed by neuroscience evidence \cite{hollerman2000involvement}, and in this work, we present an algorithm that mimics this process. Our approach couples a deep learning generative subgoal modeling with classical search algorithms to allow for successful planning with subgoals. We showcase the efficiency of our method on the following complex reasoning tasks: two popular puzzle games Sokoban and the Rubik's Cube, and an inequality theorem proving benchmark INT \cite{wu2020int}, achieving the state-of-the-art results in INT and competitive results for the remaining two. 

The deep learning revolution has brought spectacular advancements in pattern recognition techniques and models. Given the hard nature of reasoning problems, these are natural candidates to provide search heuristics \cite{bengio2020machine}. Indeed, such a blend can produce impressive results \cite{Silver2017MasteringCA,Silver2017,polu2020generative,Agostinelli:2019aa}. These approaches seek solutions using elementary actions. Others, e.g. \cite{nair2019hierarchical,DBLP:conf/nips/PertschREZJFL20, DBLP:conf/nips/KimAB19}, utilize variational subgoals generators to deal with long-horizon visual tasks. We show that these ideas can be pushed further to provide algorithms capable of dealing with combinatorial complexity. 

We present \method{} (\abbrv{}) method and give its practical implementations: MCTS-\abbrv{} and  BF-\abbrv{}. \abbrv{} consists of the following four components: 
planner, subgoal generator, a low-level policy, and a value function. 
The planner is used to search over the graph induced by the subgoal generator and is guided by the value function. The role of the low-level policy is to prune the search tree as well as to transition between subgoals.
In this paper, we assume that the generator predicts subgoals that are $k$ step ahead (towards the solution) from the current state, and to emphasize this we henceforth add $k$ to the method's abbreviation. 
MCTS-\abbrv{} and BF-\abbrv{} differ in the choice of the search engine: the former uses Monte-Carlo Tree Search (MCTS), while the latter is backed by Best-First Search (BestFS). 
We provide two sets of implementations for the generator, the low-level policy, and the value functions. The first one uses transformer architecture \cite{vaswani2017attention} for each component, while the second 
utilizes a convolutional network for the generator and the value function, and the classical breadth-first search for the low-level policy. This lets us showcase the versatility and effectiveness of the approach. 

The subgoal generator lies at the very heart of \method{}, being an implementation of reasoning with high-level ideas. To be useful in a broad spectrum of contexts, the generator should be implemented as a learnable (generative) model. As a result, it is expected to be imperfect and (sometimes) generate incorrect predictions, which may turn the search procedure invalid.  Can we thus make planning with learned subgoals work? In this paper, we answer this question affirmatively: we show that the autoregressive framework of transformer-based neural network architecture \cite{DBLP:conf/nips/VaswaniSPUJGKP17} leads to superior results in challenging domains.    

We train the transformer with the objective to predict the $k$-th step ahead. The main advantages of this subgoal objective are simplicity and empirical efficiency. We used expert data to generate labels for supervised training. When offline datasets are available, which is the case for the environments considered in this paper\footnotemark{}, such an approach allows for stable and efficient optimization with high-quality gradients. \footnotetext{The dataset for INT or Sokoban can be easily generated or are publicly available. For the Rubik's Cube, we use random data or simple heuristic (random data are often sufficient for robotic tasks and navigation.)} Consequently, this method is often taken when dealing with complex domains (see e.g. \cite{silver2016mastering, Vinyals2019}) or when only an offline expert is available\footnote{For example, the INT engine can easily generate multiple proves of random statements, but \textit{cannot} prove a given theorem.}.   Furthermore, we found evidence of out-of-distribution generalization.

Finally, we formulate the following hypothesis aiming to shed some light on why \abbrv{} is successful: we speculate that subgoal generation may alleviate errors in the value function estimation. Planning methods based on learning, including \abbrv{}, typically use imperfect value function-based information to guide the search. While traditional low-level search methods are susceptible to local noise, subgoal generation allows for evaluations of the value functions at temporally distant subgoals, which improves the signal-to-noise ratio and allows a ``leap over'' the noise. 

To sum up, our contributions are: 
\begin{enumerate} 
    
    \item We propose \method{} method with two implementations: MCTS-\abbrv{}, BF-\abbrv{}. We demonstrate that our approach requires a relatively little search or, equivalently, is able to handle bigger problems. We also observe evidence of out-of-distribution generalization.
    
    
    \item  We provide evidence that a transformer-based autoregressive model learned with a simple supervised objective to predict states $k$-th step ahead is an effective tool to generate valid and diverse subgoals.
    
    
    \item We show in our experiments that using subgoal planning help to might mitigate the negative influence of value function errors on planning.
    
\end{enumerate}

We provide the code of our method and experiment settings at \url{https://github.com/subgoal-search/subgoal-search}, and a dedicated website \url{https://sites.google.com/view/subgoal-search}.

\section{Related work}

In classical AI, reasoning is often achieved by \textit{search} (\cite{russell2002artificial}). Search rarely can be exhaustive, and a large body of algorithms and heuristics has been developed over the years, \cite[Section 3.5]{russell2002artificial}. It is hypothesized that progress can be achieved by combining search with learning \cite{bengio2020machine}. Among notable successful examples of this approach are
Alpha Zero \cite{Silver2018}, or solving Rubik's cube using a learned heuristic function to guide the $\text{A}^*$ algorithm (see \cite{Agostinelli:2019aa}).

An eminent early example of using goal-directed reasoning is the PARADISE algorithm (\cite{DBLP:journals/ai/Wilkins80}). In deep learning literature, \cite{DBLP:conf/nips/KurutachTYRA18} was perhaps the first work implementing subgoal planning. This was followed by a large body of work on  
planning with subgoals in the latent space for visual tasks  (\cite{DBLP:conf/nips/KimAB19, DBLP:conf/nips/NasirianyPLL19, DBLP:conf/nips/PertschREZJFL20, nair2019hierarchical, DBLP:conf/iclr/JayaramanEEL19, fang2019dynamics, DBLP:conf/l4dc/PertschRYZDDLJ20}) 
or landmark-based navigation methods (\cite{SavinovDK18, DBLP:conf/icml/LiuKTAT20, DBLP:conf/corl/GaoHLSS17, DBLP:conf/corl/SteinBR18, DBLP:journals/corr/abs-2011-12491}). 

The tasks considered in the aforementioned studies are often quite forgiving when it comes to small errors in the subgoal generation. This can be contrasted with complex reasoning domains, in which even a small variation of a state may drastically change its meaning or render it invalid. Thus, neural networks may struggle to generate semantically valid states (\cite[Section 6.1]{bengio2020machine}).

Assuming that this problem was solved, a generated subgoal still remains to be assessed. The exact evaluation may, in general, require exhaustive search or access to an oracle (in which case the original problem is essentially solved). Consequently, it is unlikely that a simple planner (e.g., one unrolling independent sequences of subgoals \cite{fang2019dynamics}) will either produce an informative outcome, could be easily improved using only local changes via gradient descent \cite{DBLP:conf/nips/NasirianyPLL19}, or cross-entropy method (CEM) \cite{nair2019hierarchical, DBLP:conf/nips/PertschREZJFL20,  DBLP:conf/l4dc/PertschRYZDDLJ20}. Existing approaches based, which are based on more powerful subgoal search methods, have their limitations, on the other hand. \cite{gabor2019subgoal} is perhaps the closest to our method and uses MCTS to search the subgoal-induced graph. However, it uses a predefined (not learned) predicate function as a subgoal generator, limiting applicability to the problems with available high-quality heuristics. Learned subgoals are used in \cite{parascandolo2020divide}, a method of hierarchical planning. That said, the subgoal space needs to be relatively small for this method to work (or crafted manually to reduce cardinality). To the best of our knowledge, our algorithm is the first domain agnostic hierarchical planner for combinatorially complex domains. 

More generally, concepts related to goals and subgoals percolated to reinforcement learning early on, leading, among others, to prominent ideas like hindsight \cite{DBLP:conf/ijcai/Kaelbling93}, hierarchical learning \cite{DBLP:journals/ai/SuttonPS99, DBLP:conf/nips/DayanH92} or the Horde architecture \cite{DBLP:conf/atal/SuttonMDDPWP11}. Recently, with the advent of \textit{deep} reinforcement learning, these ideas have been resurfacing and scaled up to deliver their initial promises. For example, \cite{Vezhnevets2017FeUdalNF} implements ideas of \cite{DBLP:conf/nips/DayanH92} and a very successful hindsight technique \cite{hind19} is already considered to be a core RL method.   Further, a recent paper \cite{DBLP:conf/icml/PitisCZSB20} utilizes a maximum entropy objective to select achievable goals in hindsight training. Apart from the aforementioned hierarchical planning, the idea to use neural networks for subgoal generation was used to improve model-free reinforcement learning agents \cite{florensa2018automatic, chane2021goal} and imitation learning algorithms \cite{paul2019learning}.

\section{Method}\label{sec:methods}
Our method, \method{} (\abbrv), is designed for problems, which can be formulated as a search over a graph with a known transition model. Formally, let $G = (\mathcal{S}, \mathcal{E})$ be a directed graph and $\tilde{\mathcal S} \subset \mathcal S$ be the set of success states. We assume that, during the solving process, the algorithm can, for a given node $g$, determine the edges starting at $g$ and check if $g \in \tilde{S}$. 

\method{} consists of four components:  planner, subgoal generator, low-level policy, and a value function. The planner, coupled with a value function, is used to search over the graph induced by the subgoal generator. Namely, for each selected subgoal, the generator allows for sampling the candidates for the next subgoals. Only these reachable by the low-level policy are used. The procedure continues until the solution is found or the computational budget is exhausted. Our method searches over a graph $\tilde{G} = (\mathcal{S}, \mathcal{E}_s)$, with the edges $\mathcal{E}_s$ given by the subgoal generator. Provided a reasonable  generator, paths to $\tilde{\mathcal{S}}$ are shorter in $\tilde{G}$ than in ${G}$ and thus easier to find.

In this paper we provide BestFS- and MCTS- backed implementations \abbrv{}. Algorithm \ref{alg:BestFS_with_subgoals} presents BF-\abbrv{}; see Algorithm \ref{alg:generic_mcts} in Appendix \ref{sec:mcts_algorithm_appendix} for MCTS-\abbrv{}.

For INT and Rubik's Cube, we use transformer models (see Appendix \ref{sec:transformer_architecture}) in all components other than the planner. For Sokoban, we use convolutional networks (see Appendix \ref{sec:appendix_sokoban_generator}). While transformers could also be used in Sokoban, we show that a simplified setup already achieves strong results. This showcases that \method{} is general enough to work with different design choices. In Section \ref{sec:search_domains} we describe datasets used train these neural models.



\noindent
\begin{minipage}[t]{.52\textwidth}
\begin{algorithm}[H]
    \caption{Best-First Subgoal Search (BF-kSubS)}
    \label{alg:BestFS_with_subgoals}
\begin{tabular}{ l c l }
    \textbf{Require: }
    & $C_1$& max number of nodes \\
    & $V$ & value function network \\
    & $\Call{Solved}{}$ & predicate of solution \\
\end{tabular}
\begin{algorithmic}
    \Function{solve}{$\mathtt{s_0}$}
        \State $\mathtt{T}$.\Call{push}{($V(\mathtt{s_0}), \mathtt{s_0})$} \Comment{$\mathtt{T}$ is priority queue}
        \State $\mathtt{paths}[\mathtt{s_0}] \gets []$\Comment{$\mathtt{paths}$ is dict of lists}
        \State $\mathtt{seen}.\Call{add}{\mathtt{s_0}}$ \Comment{$\mathtt{seen}$ is set}
        \While{$0 < \Call{{len}}{\mathtt{T}} \text{ and } \Call{{len}}{\mathtt{seen}}<C_1$} 
            \State $\_, \mathtt{s} \gets \mathtt{T}.\Call{extract\_max}{ }$
            \State $\mathtt{subgoals} \gets \Call{sub\_generate}{\mathtt{s}}$
            \State \Comment{see Algorithm \ref{alg:subgoal_generator}}
            \For{$\mathtt{s'} \textbf{ in } \mathtt{subgoals}$}
                \State \textbf{if} $\mathtt{s'} \textbf{ in } \mathtt{seen}$ \textbf{then} $\mathtt{continue}$
                \State $\mathtt{seen}.\Call{add}{\mathtt{s'}}$
                \State $\mathtt{actions} \gets \Call{get\_path}{\mathtt{s}, \mathtt{s'}}$
                \State \Comment{see Alg \ref{alg:conditional_policy} or Alg \ref{alg:sokoban_conditional_policy}}
                \If{$\mathtt{actions}.\Call{empty}{ }$}
                    \State $\mathtt{continue}$
                \EndIf
                \State $\mathtt{T}$.\Call{push}{($V(\mathtt{s'}), \mathtt{s'})$}
                \State $\mathtt{paths}[\mathtt{s'}] \gets \mathtt{paths}[\mathtt{s}]+\mathtt{actions}$
                \If{$\Call{solved}{\mathtt{s'}}$}
                    \State \Return $\mathtt{paths}[\mathtt{s'}]$
                \EndIf
            \EndFor
        \EndWhile
        \State \Return $\mathtt{False}$
    \EndFunction
\end{algorithmic}
\end{algorithm}
\end{minipage}
\hfill
\noindent
\begin{minipage}[t]{.48\textwidth}
\begin{algorithm}[H]
    \caption{Low-level conditional policy}
    \label{alg:conditional_policy}
\begin{tabular}{ l c l }
    \textbf{Require: }
    & $C_2$ & steps limit \\
    & $\pi$ & low-level conditional\\
    & & policy network\\
    & $M$ & model of the environment \\
\end{tabular}
\begin{algorithmic}
    \Function{get\_path}{$\mathtt{s_0}$, $\mathtt{subgoal}$}
        \State $\mathtt{step} \gets 0$
        \State $\mathtt{s} \gets \mathtt{s_0} $
        \State $\mathtt{action\_path} \gets []$
        \While{$\mathtt{step} < C_2 $ }
            \State $\mathtt{action} \gets \pi.\Call{predict}{\mathtt{s},\mathtt{subgoal}}$
            \State $\mathtt{action\_path}.\Call{append}{\mathtt{action}}$
            \State $\mathtt{s} \gets M.\Call{next\_state}{\mathtt{s, action}}$

            \If{$\mathtt{s} = \mathtt{subgoal}$}
                \State \Return $\mathtt{action\_path}$
            \EndIf
            \State $\mathtt{step} \gets \mathtt{step} + 1$
        \EndWhile
        \State \Return $[]$
    \EndFunction
\end{algorithmic}
\end{algorithm}
\end{minipage}

\textbf{Subgoal generator} Formally, it is  a mapping $\rho\colon\mathcal{S} \to \mathbf{P}(\mathcal{S})$, where $\mathbf{P}(\mathcal{S})$ is a family of probability distributions over the environment's state space $\mathcal{S}$. More precisely, let us define a trajectory as a sequence of state-action pairs $(s_0, a_0), (s_1, a_1), \dots, (s_n, a_n)$, with $(s_i, a_i)\in \mathcal{S}\times\mathcal{A}$, where $\mathcal{A}$ stands for the action space and $n$ is the trajectory's length. We will say that the generator predicts $k$-\textit{step ahead} subgoals, if at any state $s_\ell$ it aims to predict $s_{\min(\ell+k, n)}$. We show, perhaps surprisingly, that this simple objective is an efficient way to improve planning, even for small values of $k$, i.e. $k\in \{2, 3, 4, 5\}$.

Operationally, the subgoal generator takes as input an element of the $\textbf{s} \in \mathcal S$ and returns \textit{subgoals}, a set of new candidate states expected to be closer to the solution and is implemented by Algorithm \ref{alg:subgoal_generator}. It works as follows: first, we generate $C_3$ subgoal candidates with \textsc{sub\_net\_generate}. Then, we prune this set to obtain a total probability greater than $C_4$. 

For INT and Rubik's Cube, we represent states as sequences modeled with a transformer. Following the practice routinely used in language modeling, \cite{vaswani2017attention}, \textsc{sub\_net\_generate} employs beam search to generate a set of high likelihood outputs.\footnote{In language modeling, typically, only one beam search output is used. In our case, however, we utilize all of them, which turns out to be a diverse set of subgoals.} With the same goal in mind, for Sokoban, we use another method described Appendix \ref{sec:appendix_datasets_sokoban}. \textsc{sub\_net\_generate} uses the subgoal generator network, which is trained in a supervised way on the expert data, with training examples being: $s_\ell$ (input) and $s_{\min(\ell+k, n)}$ (label), see Appendix \ref{sec:training_details_appendix} for details.

\begin{wrapfigure}{R}{0.6\textwidth}
\vspace{-0.7cm}
\begin{minipage}[t]{0.6\textwidth}
\begin{algorithm}[H]
    \caption{Subgoal generator}
    \label{alg:subgoal_generator}
\begin{tabular}{ l c l }
    \textbf{Require: }
    & $C_3$ & number of subgoals \\
    & $C_4$ & target probability \\
    & $\rho$ & subgoal generator network \\ 
\end{tabular}
\begin{algorithmic}
    \Function{sub\_generate}{$\mathtt{s}$}
        \State $\mathtt{subgoals} \gets \emptyset$
        
        \State $\mathtt{states}, \mathtt{probs} \gets 
        \Call{sub\_net\_generate}{\rho,\mathtt{s}; C_3}$ 
        \State \Comment{$(\mathtt{states}, \mathtt{probs})$ is sorted wrt $\mathtt{probs}$ }
        \State $\mathtt{total\_p} \gets 0$
        \For{$\mathtt{state}, \mathtt{p} \in (\mathtt{states}, \mathtt{probs})$}
            \State \textbf{if} {$\mathtt{total\_p} > C_4$} \textbf{ then break}
            \State $\mathtt{subgoals}.\Call{add}{\mathtt{state}}$
            \State $\mathtt{total\_p} \gets \mathtt{total\_p} + \mathtt{p}$
        \EndFor
        \State \Return $\mathtt{subgoals}$
    \EndFunction
\end{algorithmic}
\end{algorithm}
\end{minipage}
\end{wrapfigure}

\textbf{Low-level conditional policy} 
Formally, it is a mapping $\pi\colon \mathcal S\times \mathcal S\to \mathcal A^*$. It is used to generate a sequence of actions on how to reach a subgoal starting from a given initial state. Operationally, it may return an empty sequence if the subgoal cannot be reached within $C_2$ steps, see Algorithm \ref{alg:conditional_policy}. This is used as a pruning mechanism for the \textit{subgoals} set in Algorithm \ref{alg:BestFS_with_subgoals}. 

In INT and Rubik's Cube, we use Algorithm \ref{alg:conditional_policy}, which utilizes low-level policy network $\pi$. Similarly to the subgoal generator, it is trained using expert data in a supervised fashion, i.e. for a pair  $(s_\ell, s_{\min(\ell+i, n)})$, with $i\le k$, its objective is to predict $a_\ell$. 

When the branching factor is small, the low-level policy can be realized by a simple breadth-first search mechanism, see Algorithm \ref{alg:sokoban_conditional_policy} in Appendix, which we illustrate on Sokoban.

\textbf{Value function} Formally, it is a mapping $V\colon \mathcal S \to \mathbb R$, that assigns to each state a value related to its distance to the solution, and it is used to guide the search (see Algorithm \ref{alg:BestFS_with_subgoals} and Algorithm \ref{alg:generic_mcts}). For its training, we use expert data. For each state $s_\ell$ the training target is negated distance to the goal: $\ell - n$, where $n$ denotes the end step of a trajectory that $s_\ell$ belongs to.

\textbf{Planner} 
This is the engine that we use to search the subgoal-induced graph. 
In this paper, we use BestFS (Algorithm \ref{alg:BestFS_with_subgoals}) and  MCTS  (Algorithm \ref{alg:generic_mcts} in Appendix \ref{sec:mcts_algorithm_appendix}). The former is a classic planning method, which maintains a priority queue of states waiting to be explored, and greedily (with respect to their value) selects elements from it (see, e.g., \cite{russell2002artificial}). MCTS is a search method that iteratively and explicitly builds a search tree, using (and updating) the collected node statistics (see, e.g., \cite{browne2012survey}). In this paper, we use an AlphaZero-like \cite{silver2016mastering} algorithm for single-player games.

We note that the subgoal generator can be combined with virtually any search algorithm and can benefit from an additional structure. For example, for domains providing a factored state representation, the width-based methods \cite{lipovetzky2012width, frances2017purely} would likely be stronger search mechanisms. 


\section{Experiments}
In this section, we empirically demonstrate the efficiency of MCTS-\abbrv{} and BF-\abbrv{}. In particular, we show that they vastly outperform their standard (``non-subgoal'') counterparts. As a testing ground, we consider three challenging domains: Sokoban, Rubik's Cube, and INT.  All of them require non-trivial reasoning. The Rubik's Cube is a well-known 3-D combination puzzle. Sokoban is a complex video puzzle game known to be NP-hard and thus challenging for planning methods. INT \cite{wu2020int} is a recent theorem proving benchmark.

\subsection{Training protocol and baselines}
Our experiments consist of three stages. First, we collect domain-specific expert data, see Section \ref{sec:search_domains}. Secondly, we train the subgoal generator, low-level conditional policy, and value function networks using the data and targets described in Section \ref{sec:methods}. For more details see Appendix \ref{sec:training_details_appendix}. Eventually, we evaluate the planning performance of MCTS-\abbrv{} and BF-\abbrv{}, details of which are presented below. In the second step, we use supervised learning, which makes our setup stable with respect to network initialization, see details in Appendix \ref{sec:seeds_appendix}. 

As baselines, we use BestFS and MCTS (being a single-player implementation of AlphaZero). In INT and Rubik's Cube, both the algorithms utilize policy networks (trained with behavioral cloning, on the same dataset, which we used to train \abbrv{}). Note that distribution over actions induces a distribution over states; thus the policy network can be regarded as a subgoal generator for $k=1$. More details about the baselines can be found in Appendix \ref{sec:baslines_app}.

\subsection{Search Domains and Datasets}\label{sec:search_domains}


\textbf{Sokoban} is a single-player complex game in which a player controls an agent whose goal is to place boxes on target locations solely by pushing them; without crossing any obstacles or walls. Sokoban has recently been used to test the boundaries in RL \cite{guez2019investigation, milos2019uncertainty}. Sokoban is known to be hard \cite{fern2011first}, mainly due to its combinatorial complexity and the existence of irreversible states. Deciding if a given Sokoban board is solvable is an NP-hard problem \cite{dor1999sokoban}.

We collect the expert dataset consisting of all successful trajectories occurring during the training of an MCTS agent (using an implementation of \cite{milos2019uncertainty}). These are suboptimal, especially in the early phases of the training or for harder boards. For both expert training and \abbrv{} evaluation, we generate Sokoban boards following the approach of \cite{RacaniereWRBGRB17}.

\textbf{Rubik's Cube} is a classical 3-dimensional puzzle. It is considered challenging due to the fact that the search space has more than $4.3 \times 10^{18}$ configurations. Similarly to Sokoban, Rubik's Cube has been recently used as a testing ground for RL methods \cite{Agostinelli:2019aa}.

To collect the expert dataset,  we generate random paths of length $30$ starting from the solved cube and take them backward. These backward solutions are highly sub-optimal (optimal solutions are proven to be shorter than $26$ \cite{rokicki2014god}).

\textbf{INT: Inequality Benchmark for Theorem Proving}. INT provides a generator of inequalities, which produces mathematical formulas along with their proofs, see \cite[Section 3.3]{wu2020int}. Proofs are represented as sequences of consecutively applied mathematical axioms (there are $K=18$ axioms in total). An action in INT is a tuple containing an axiom and its input entities. The action space in this problem can be very large, reaching up to $10^6$ elements, which significantly complicates planning. 

\begin{wraptable}{R}{0.35\textwidth}
\begin{tabular}{lrrr}
\toprule
                         & INT & Sokoban & Rubik \\
\midrule
$k$    & 3   & 4       & 4     \\
$C_1$    & 400 & 5000    & 1500  \\
$C_2$  & 4   & 4       & 7     \\
$C_3$  & 4   & 4       & 3     \\
$C_4$   & 1   & 0.98    & 1 \\
\bottomrule
\end{tabular}
\caption{\small BF-\abbrv{} hyperparameters.} 
\label{tab:BFSubS_hyperparams}
\end{wraptable}

The INT generator constructs paths by randomly applying axioms starting with a trivial statement. Such a path taken backward constitutes the proof of its final statement (not guaranteed to be optimal). The proof length, denoted by $L$, is an important hyperparameter regulating the difficulty -- we use $5, 10, 15$. 

For more details on datasets see Appendix \ref{sec:data_processing_appendix}.

\subsection{Main results}\label{sec:main_results}

In this section, we present our most important finding: \method{} enables for more efficient search and consequently scales up to problems with higher difficulty. Specifically,  MCTS-\abbrv{} and BF-\abbrv{} perform significantly better than the respective methods not using subgoals, including state-of-the-art on INT. 

\begin{figure}
\centering \small
\begin{minipage}[t]{.49\textwidth}
\centering
\includegraphics[width=0.9\textwidth]{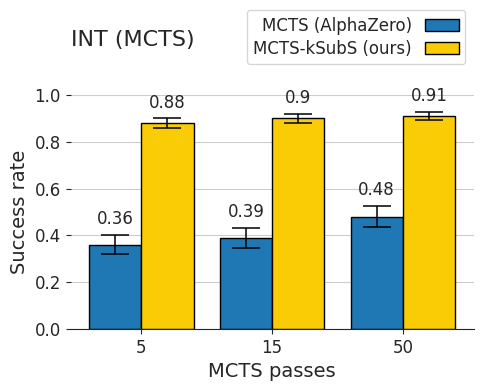}
\label{fig:int_main_results}
\end{minipage}%
\begin{minipage}[t]{0.49\textwidth}
    \centering 

\includegraphics[width=0.9\textwidth]{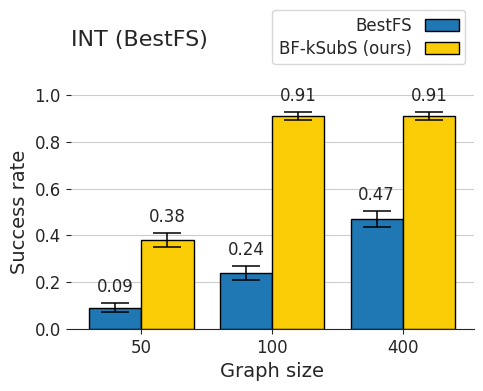}
\end{minipage}%
\vspace{0.5cm}
\begin{minipage}[t]{0.49\textwidth}
    \centering 
\includegraphics[width=0.9\textwidth]{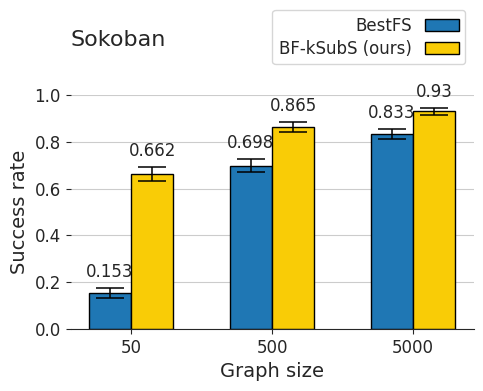}
\end{minipage}
\begin{minipage}[t]{0.49\textwidth}
    \centering 
\includegraphics[width=0.9\textwidth]{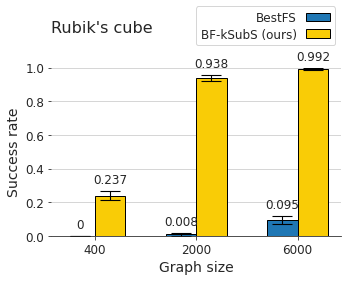}
\end{minipage}
\caption{\small The performance of \method{}. (top, left) comparison on INT (with the proof length 15) to AlphaZero. (top, right) BF-\abbrv{} consistently achieves high performance even for small computational budgets. (bottom, left) similarly on Sokoban (board size 12x12 with 4 boxes) the advantage of BF-\abbrv{} is clearly visible for small budget. (bottom, right) BestFS fails to solve Rubik's Cube, while BF-\abbrv{} can achieve near-perfect performance. }\label{fig:main_results}
\end{figure}

In Figure \ref{fig:main_results}, we present the performance of \method{}. We measure the \textit{success rate} as a function of the \textit{search budget}. The success rate is measured on $1000$ instances of a given problem (which results in confidence intervals within $\pm 0.03$). For BF-\abbrv{} the search budget is referred to as \textit{graph size} and includes the number of nodes visited by Algorithm \ref{alg:BestFS_with_subgoals}. For INT and Rubik's Cube, we include both the subgoal generated by \texttt{SUB\_GENERATE} and the nodes visited by \texttt{GET\_PATH} (as they induce a significant computational cost stemming from using low-level policy $\pi$ in Algorithm \ref{alg:conditional_policy}). For Sokoban, we use Algorithm \ref{alg:sokoban_conditional_policy} to realize \texttt{GET\_PATH}, as it has a negligible cost (less than $1\%$ of the total runtime of Algorithm 1), we do not include these nodes into graph size.

For MCTS, we report \textit{MCTS passes}, which is a common metric for MCTS, see details in Appendix \ref{sec:mcts_algorithm_appendix}.

Below we discuss the results separately for each domain. We provide examples of solutions and generated subgoals in Appendix \ref{sec:example_subgoals_appendix}.

\textbf{INT} The difficulty of the problems in INT increases fast with the proof length $L$ and the number of accessible axioms. W used $K=18$; all of available axioms. We observe, that BF-\abbrv{} scales to proofs of length $L=10$ and $L=15$, which are significantly harder than $L=5$ considered in \cite{wu2020int}, see Table \ref{table:int_success_rates}. The same holds for MCTS-\abbrv{}, see Appendix \ref{sec:detailed_results_mcts_appendix}. 

\begin{table}[h]
\begin{tabular}{cl|cc|cc|cc}
\toprule
&Proof length           & \multicolumn{2}{|c|}{5} & \multicolumn{2}{c|}{10} & \multicolumn{2}{c}{15}                 \\ 
\midrule
&Method             & 
{\small BestFS}   & {{\small  BF-kSubS {\scriptsize(ours)}}}   & {\small BestFS}     & {{\small BF-kSubS {\scriptsize(ours)}}}   & {\small BestFS}  & {{\small BF-kSubS {\scriptsize(ours)}}}  \\ 
\midrule
\parbox[t]{2mm}{\multirow{3}{*}{\rotatebox[origin=c]{90}{Graph size}}} &50                     & 0.82      & \textbf{0.99}       & 0.47       & \textbf{0.97}       & 0.09    & \textbf{0.38}    \\ 
&100                    & 0.89      & \textbf{0.99}       & 0.64       & \textbf{0.99}       & 0.24    & \textbf{0.91}      \\ 
&200                    & 0.92      & \textbf{0.99}       & 0.67       & \textbf{0.99}       & 0.35    & \textbf{0.91}        \\ 
&400                    & 0.93      & \textbf{0.99}       & 0.72       & \textbf{0.99}       & 0.47    & \textbf{0.91}      \\ 
\bottomrule
\end{tabular}

\caption{\small INT success rates for various proof lengths and graphs sizes.}
\label{table:int_success_rates}
\end{table}

We check also that MCTS-\abbrv{} vastly outperforms the baseline - AlphaZero algorithm, see Figure~\ref{fig:main_results} (top, left). An MCTS-based agent was also evaluated in \cite{wu2020int}. Its implementation uses graph neural networks architectures and achieves $92$\% success rate for $L=5$.  Our transformed-based baseline is stronger - it solves over $99$\% instances on the same problem set.

\textbf{Sokoban} Using BF-\abbrv{} allows for significantly higher success rates rates within the same computational budget, see Table~\ref{table:sokoban_different_board_sizes}. Our solution scales well to the board size as big as $20 \times 20$; note that $10 \times 10$ boards are typically used in deep RL research \cite{guez2019investigation,RacaniereWRBGRB17}. Importantly, we observe that already for a small computational budget (graph size 1000) BF-\abbrv{} obtains higher success rates than the expert we used to create the dataset (these are $78$\%, $67$\%, $60$\% for the respective board sizes). 

We also tested how the quality of BF-\abbrv{} depends on the size of the training dataset for Sokoban, the results can be found in Appendix \ref{section:dataset_sizes}.

\begin{table}[h]
  \centering
 \begin{tabular}{cl|cc|cc|cc}
 \toprule
& Board size & \multicolumn{2}{c|}{12 x 12} & \multicolumn{2}{c|}{16 x 16} & \multicolumn{2}{c}{20 x 20} \\ \midrule
& Method             & 
{\small BestFS}   & {{\small  BF-kSubS {\scriptsize(ours)}}}   & {\small BestFS}     & {{\small BF-kSubS {\scriptsize(ours)}}}   & {\small BestFS}  & {{\small BF-kSubS {\scriptsize(ours)}}}  \\ 
\midrule
\parbox[t]{2mm}{\multirow{3}{*}{\rotatebox[origin=c]{90}{Graph size}}}
 & 50         & 0.15         & \textbf{0.66}          & 0.04         & \textbf{0.42}          & 0.02         & \textbf{0.46}          \\ 
& 100        & 0.46         & \textbf{0.79}          & 0.23         & \textbf{0.62}          & 0.10         & \textbf{0.55}          \\ 
& 1000       & 0.75         & \textbf{0.89}          & 0.61         & \textbf{0.79}          & 0.47         & \textbf{0.70}          \\ 
& 5000       & 0.83         & \textbf{0.93}          & 0.69         & \textbf{0.85}          & 0.58         & \textbf{0.77}          \\ 
\bottomrule
  \end{tabular}
  \caption{\small Sokoban success rates for various board sizes (each with 4 boxes). }
  \label{table:sokoban_different_board_sizes}
  \end{table}

\textbf{Rubik's Cube} BF-\abbrv{} solves nearly $100$\% of cubes, BestFS solve less than $10$\%, see Figure \ref{fig:main_results} (bottom, right). This is perhaps the most striking example of the advantage of using a subgoal generator instead of low-level actions. We present possible explanation in Appendix \ref{sec:appendix_rubik_baseline}.

\begin{figure}[!htb]
    \centering
    \begin{minipage}{.45\textwidth}
        \centering
        \includegraphics[width=0.95\linewidth]{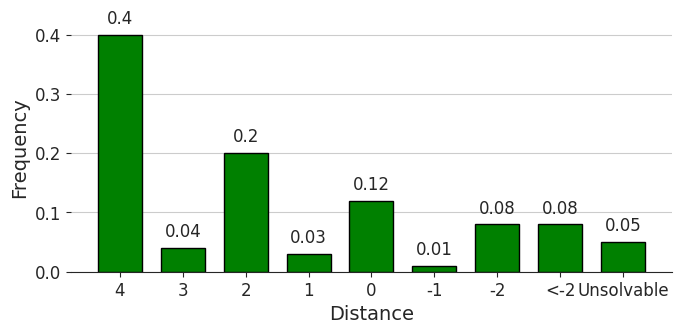}
        \caption{\small Histogram of $\Delta$. 
    Note that $17$\% of subgoals increases the distance. Additional, $5$\% leads to unsolvable ``dead states'' present in Sokoban.}
    \label{fig:distance_histogram}
  
\label{fig:sokoban}
    \end{minipage}%
    \hspace{0.5cm}
    \begin{minipage}{0.45\textwidth}
       \centering \includegraphics[width=0.95\linewidth]{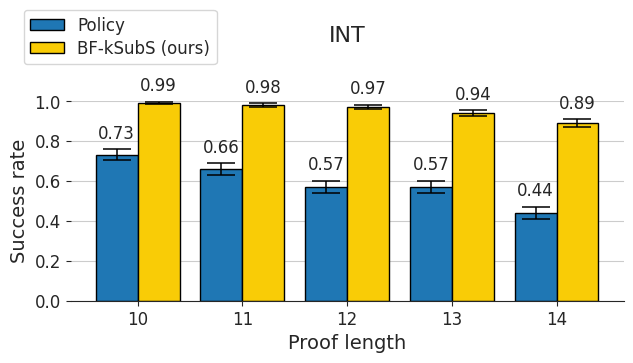}
      \caption{\small Out-of-distribution generalization to longer proofs. We compare with the behavioral cloning agent (Policy) studied in \cite{wu2020int}.}
          
	\label{fig:int_generalization}
    \end{minipage}
\end{figure}

\textbf{Out-of-distribution (OOD) generalization} OOD generalization is considered to be the crucial ability to make progress in hard combinatorial optimization problems \cite{bengio2020machine} and automated theorem proving \cite{wu2020int}. The INT inequality generator has been specifically designed to benchmark this phenomenon. We check that \method{} trained on proofs on length $10$ generalizes favourably to longer problems, see Figure~\ref{fig:int_generalization}. Following \cite{wu2020int}, we speculate that search is a computational mechanism that delivers OOD generalization.

It might be hard to compare computational budgets between various algorithms and their versions. In Appendix \ref{sec:wall_time_appendix} we measure that BF-\abbrv{} and MCTS-\abbrv{} offer very practical benefits, sometimes as much as $7\times$ faster execution. 


\subsection{Analysis of $k$ (subgoal distance) parameter}\label{sec:subgoal_distance}

The subgoals are trained to predict states $k$ steps ahead of the current one. Higher $k$ should make planning easier as the search graph is smaller. However, as $k$ increases, the quality of the generator may drop, and thus the overall effect is uncertain. Similarly, the task of the low-level conditional policy becomes more difficult as $k$ increases. The optimal value of $k$ is $3$ and $4$ for INT and Rubik's Cube, respectively. In these environments, increasing $k$ further degrades performance. In Sokoban, we observe monotonic improvement up to $k=10$. This is perhaps because low-level conditional policy (Algorithm \ref{alg:sokoban_conditional_policy}, based on breadth-first search) never fails to fill the path from a state to the valid subgoal. The running cost of Algorithm \ref{alg:sokoban_conditional_policy} quickly becomes unacceptable (recall that for $k=4$, which we used in the main experiment, it has still a negligible cost - below $<1\%$ of the total runtime).



\begin{figure}[h]
\centering \small
\begin{minipage}[t]{.33\textwidth}
\centering
\includegraphics[width=1.\textwidth]{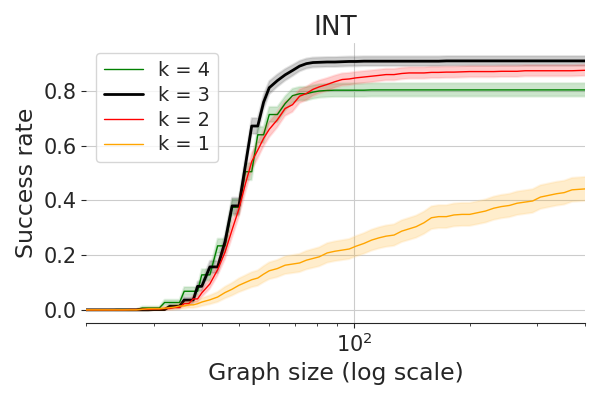}
\label{fig:int_main_results}
\end{minipage}%
\begin{minipage}[t]{0.33\textwidth}
    \centering 

\includegraphics[width=1.\textwidth]{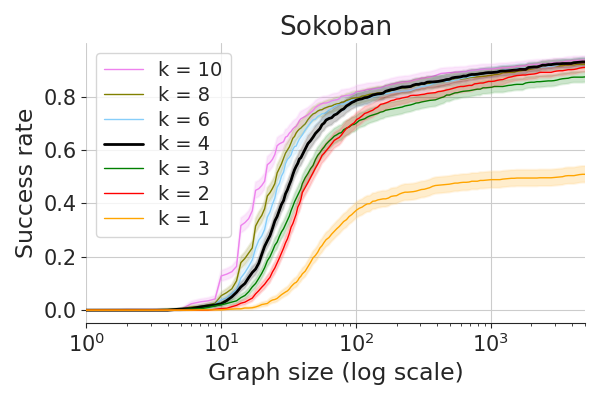}
\end{minipage}%
\vspace{0.5cm}
\begin{minipage}[t]{0.33\textwidth}
    \centering 
\includegraphics[width=1.\textwidth]{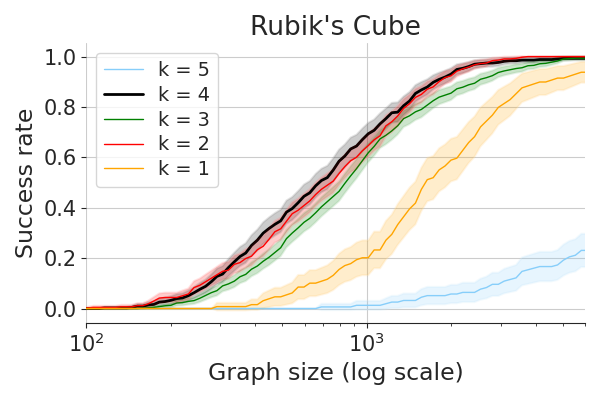}
\end{minipage}
\caption{\small BF-\abbrv{} success rates for different values of $k$. Black curves represent the values of $k$ used in the main experiments (that is $k=4$ for Rubik's Cube and Sokoban and $k=3$ for INT).  }\label{fig:different_k}
\end{figure}

\subsection{Quality of subgoals}\label{sec:subgoals_closer_look}

The learned subgoal generator is likely to be imperfect (especially in hard problems). We study this on $10\times 10$ boards of Sokoban, which are small enough to calculate the true distance $dist$ to the solution using the Dijkstra algorithm. In Figure \ref{fig:distance_histogram}, we study $\Delta := dist_{s_1} - dist_{s_2}$, where $s_1$ is a sampled state and $s_2$ is a subgoal generated from $s_1$. Ideally, the histogram should concentrate on $k=4$ used in training. We see that in slightly more than $65\%$ of cases subgoals lead to an improvement.

The low-level conditional policy in Algorithm \ref{alg:conditional_policy} provides additional verification of generated states.  We check that in INT and Rubik's Cube, about $50$\% of generated subgoals can be reached by this policy (the rest is discarded). 

\subsection{Value errors}\label{sec:local_guidance}
There might be various explanations for the success of our method. One of them is that \method{} better handles errors of learned value functions. In this section, we discuss this using a synthetic grid world example and performing statistical analysis on the real value function trained to approximate the distance to the solution (as described in Section \ref{sec:methods}).

\textbf{Grid world example}
\begin{wraptable}{R}{0.30\textwidth}
\begin{tabular}{l|cc}
\toprule
                  $\sigma$       & BestFS & BF-\abbrv{} \\
\midrule
$3$    & 0.999   & 1       \\
$10$    & 0.142 & 1  \\
$20$  & 0.006   & 0.983       \\
\bottomrule
\end{tabular}
\caption{\small Success rates on the grid world ($m=6, n=10$), depending on the value function noise scale. We use the search budget of $500$ nodes and  $k=4$ for \abbrv{}.} 
\vspace{-0.6cm}
\label{tab:grid_world}
\end{wraptable}
Consider a grid world with the state space $S=$\{1, \ldots, n\}$^m$, with $(0, \dots, 0)$ being the initial state and $(n, \dots, n)$ the goal state. A pair of states is connected by an edge if they are at distance $1$ from each other. Let:

\begin{itemize}
    \item Synthetic value function: the negative distance to the goal plus i.i.d Gaussian noise $\mathcal{N}(0, \sigma^2$).
    \item Synthetic $\Call{sub\_generate}{}$ (instead of Algorithm \ref{alg:subgoal_generator}): Let $B_k(s)$ be the states within distance $k$ from $s$. We return $C_3 - 1$ states sampled uniformly from $B_k(s)$ and a one "good subgoal" being a state in $B_k(s)$ with the minimal distance to the solution.
    \item Node expansion in BestFS (baseline): implemented as $\Call{sub\_generate}{}$ above with  $k = 1$.
\end{itemize}

In this setup, one easily sees that the probability that the good subgoal will have the highest value estimation among the generated states grows with $k$. Consequently,  \abbrv{} can handle higher levels of noise than the baseline BestFS, see Table~\ref{tab:grid_world}.

 \textbf{Value monotonicity} Imagine a solution path from the starting state to the goal state. Due to the errors, the value estimates on the path may not be monotonic. This is an undesirable property, which is likely to hinder search and make finding the path harder. Now consider the subpath consisting of consecutive states spaced $k$ actions apart, as can be constructed by \method{}. For this version, the value estimates are more likely to be monotonic and easier to find. To illustrate this, we measure monotonicity on solution paths found by our algorithm for INT. The probability that value decreases when moving $k$ steps ahead drops from $0.32$ when $k = 1$ to mere $0.02$ for $k = 4$ (see Table \ref{table:value_decrease} in Appendix).

\textbf{Overoptimism} Alternatively, one can consider that erroneously positive values misguide a search method (a phenomenon known as over-optimism \cite{hasselt2010double}). To illustrate this, consider $\mathcal S_3(s)$, the set of all states 
having the same distance to the solution as~$s$ and within distance $3$ from $s$. Intuitively, $\mathcal S_3(s)$ contains similar states with respect to the difficulty of solving. In Sokoban, the standard deviation of value function prediction for $\mathcal S_3(s)$ is equal to $2.43$ (averaged over different $s$ on Sokoban boards). This is high when compared to the average increase of value for moving one step closer to the solution, which is only $1.34$. Consequently, it is likely that $\mathcal S_3(s)$ contains a suboptimal state, e.g., having a higher value than the best immediate neighbors of $s$ (which by properties of the game will be closer to solution in Sokoban). Indeed, we measure that the probability of such an event is $64$\%. However, it drops significantly to $29$\% if one considers states closer by $4$ steps (say given by a subgoal generator). 

\section{Limitations and future work} \label{sec:limitations}
In this section, we list some limitations of our work and suggest further research directions. 

\textbf{Reliance on expert data} In this version, we use expert data to train learnable models. As \abbrv{} improves the performance, we speculate that training akin to AlphaZero can be used, i.e. in a planner-learner loop without any outside knowledge. 

\textbf{Optimality and completeness} \abbrv{} searches over a reduced state space, which might produce suboptimal solutions or even fail to find them. This is arguably unavoidable if we seek an efficient method for complex problems. 

\textbf{Subgoals definition} We use simple $k$-step ahead subgoals, which is perhaps not always optimal. Our method can be coupled with other subgoal paradigms. Unsupervised detection of landmarks (see e.g. \cite{DBLP:journals/corr/abs-2011-12491}) seems an attractive future research direction.

\textbf{More environments} In future work, we plan to test \abbrv{} on more environments to understand its strengths and weaknesses better. In this work, we generate subgoals in the state space, which might be limiting for tasks with high dimensional input (e.g., visual).

\textbf{Reliance on a model of the environment} We use a perfect model of the environment, which is a common practice for some environments, e.g., INT. Extending \abbrv{} to use learned (imperfect) models is an important future research direction. 

\textbf{Determinism} Our method requires the environment to be deterministic. 

\textbf{OOD generalization} A promising future direction is to investigate and leverage the out-of-distribution generalization delivered by our method and compare to (somewhat contradictory) findings of \cite{DBLP:conf/iclr/HamrickFBGVWABV21, wu2020int}.

\textbf{Classical planning methods} For many search problems, the state space can be represented in factored fashion (or such representation can be learned \cite{asai2018classical}). In such cases, the search can be greatly improved with width-based methods \cite{lipovetzky2012width, frances2017purely}. It is an interesting research direction to combine \abbrv{} with such methods.

\section{Conclusions}
We propose \method{}, a search algorithm based on subgoal generator. We present two practical implementations  MCTS-\abbrv{} and  BF-\abbrv{} meant to be effective in complex domains requiring reasoning. We confirm that indeed our implementations excel in Sokoban, Rubik's Cube, and inequality benchmark INT. Interestingly, a simple $k$ step ahead mechanism of generating subgoals backed up by transformer-based architectures performs surprisingly well. This evidence, let us hypothesize, that our methods (and related) can be further scaled up to even harder reasoning tasks. 



\begin{ack}
The work of Konrad Czechowski, Tomasz Odrzygóźdź and Piotr Miłoś was supported by the Polish National Science Center grant UMO-2017/26/E/ST6/00622. The work of Piotr Miłoś and Michal Zawalski was supported by by the Polish National Science Center grant UMO-2019/35/O/ST6/03464.  We gratefully acknowledge Polish high-performance computing infrastructure PLGrid (HPC Centers: ACK Cyfronet AGH, PCSS) for providing computer facilities and support within computational grant no. PLG/2020/013619. Our experiments were managed using \url{https://neptune.ai}. We would like to thank the Neptune team for providing us access to the team version and technical support.
\end{ack}

\bibliography{bibliography}

\begin{thebibliography}{10}

\bibitem{Agostinelli:2019aa}
Forest Agostinelli, Stephen McAleer, Alexander Shmakov, and Pierre Baldi.
\newblock Solving the rubik's cube with deep reinforcement learning and search.
\newblock {\em Nature Machine Intelligence}, 1(8):356--363, 2019.

\bibitem{hind19}
Marcin Andrychowicz, Dwight Crow, Alex Ray, Jonas Schneider, Rachel Fong, Peter
  Welinder, Bob McGrew, Josh Tobin, Pieter Abbeel, and Wojciech Zaremba.
\newblock Hindsight experience replay.
\newblock In {\em Advances in Neural Information Processing Systems 30: Annual
  Conference on Neural Information Processing Systems 2017, 4-9 December 2017,
  Long Beach, CA, {USA}}, pages 5048--5058, 2017.

\bibitem{asai2018classical}
Masataro Asai and Alex Fukunaga.
\newblock Classical planning in deep latent space: Bridging the
  subsymbolic-symbolic boundary.
\newblock In {\em Thirty-Second AAAI Conference on Artificial Intelligence},
  2018.

\bibitem{bengio2020machine}
Yoshua Bengio, Andrea Lodi, and Antoine Prouvost.
\newblock Machine learning for combinatorial optimization: a methodological
  tour d’horizon.
\newblock {\em European Journal of Operational Research}, 2020.

\bibitem{browne2012survey}
Cameron~B Browne, Edward Powley, Daniel Whitehouse, Simon~M Lucas, Peter~I
  Cowling, Philipp Rohlfshagen, Stephen Tavener, Diego Perez, Spyridon
  Samothrakis, and Simon Colton.
\newblock A survey of monte carlo tree search methods.
\newblock {\em IEEE Transactions on Computational Intelligence and AI in
  games}, 4(1):1--43, 2012.

\bibitem{chane2021goal}
Elliot Chane-Sane, Cordelia Schmid, and Ivan Laptev.
\newblock Goal-conditioned reinforcement learning with imagined subgoals.
\newblock In {\em International Conference on Machine Learning}, pages
  1430--1440. PMLR, 2021.

\bibitem{DBLP:conf/nips/DayanH92}
Peter Dayan and Geoffrey~E. Hinton.
\newblock Feudal reinforcement learning.
\newblock In Stephen~Jose Hanson, Jack~D. Cowan, and C.~Lee Giles, editors,
  {\em Advances in Neural Information Processing Systems 5, {[NIPS} Conference,
  Denver, Colorado, USA, November 30 - December 3, 1992]}, pages 271--278.
  Morgan Kaufmann, 1992.

\bibitem{dor1999sokoban}
Dorit Dor and Uri Zwick.
\newblock Sokoban and other motion planning problems.
\newblock {\em Computational Geometry}, 13(4):215--228, 1999.

\bibitem{fang2019dynamics}
Kuan Fang, Yuke Zhu, Animesh Garg, Silvio Savarese, and Li~Fei-Fei.
\newblock Dynamics learning with cascaded variational inference for multi-step
  manipulation.
\newblock {\em arXiv preprint arXiv:1910.13395}, 2019.

\bibitem{fern2011first}
Alan Fern, Roni Khardon, and Prasad Tadepalli.
\newblock The first learning track of the international planning competition.
\newblock {\em Machine Learning}, 84(1-2):81--107, 2011.

\bibitem{florensa2018automatic}
Carlos Florensa, David Held, Xinyang Geng, and Pieter Abbeel.
\newblock Automatic goal generation for reinforcement learning agents.
\newblock In {\em International conference on machine learning}, pages
  1515--1528. PMLR, 2018.

\bibitem{frances2017purely}
Guillem Frances, Miquel Ram{\'\i}rez~J{\'a}vega, Nir Lipovetzky, and Hector
  Geffner.
\newblock Purely declarative action descriptions are overrated: Classical
  planning with simulators.
\newblock In {\em IJCAI 2017. Twenty-Sixth International Joint Conference on
  Artificial Intelligence; 2017 Aug 19-25; Melbourne, Australia.[California]:
  IJCAI; 2017. p. 4294-301.} International Joint Conferences on Artificial
  Intelligence Organization (IJCAI), 2017.

\bibitem{gabor2019subgoal}
Thomas Gabor, Jan Peter, Thomy Phan, Christian Meyer, and Claudia
  Linnhoff-Popien.
\newblock Subgoal-based temporal abstraction in {Monte-Carlo Tree Search}.
\newblock In {\em IJCAI}, pages 5562--5568, 2019.

\bibitem{DBLP:conf/corl/GaoHLSS17}
Wei Gao, David Hsu, Wee~Sun Lee, Shengmei Shen, and Karthikk Subramanian.
\newblock Intention-net: Integrating planning and deep learning for
  goal-directed autonomous navigation.
\newblock In {\em 1st Annual Conference on Robot Learning, CoRL 2017, Mountain
  View, California, USA, November 13-15, 2017, Proceedings}, volume~78 of {\em
  Proceedings of Machine Learning Research}, pages 185--194. {PMLR}, 2017.

\bibitem{gowers2000importance}
Timothy Gowers.
\newblock {\em The importance of mathematics}.
\newblock Springer-Verlag, 2000.

\bibitem{guez2019investigation}
Arthur Guez, Mehdi Mirza, Karol Gregor, Rishabh Kabra, S{\'e}bastien
  Racani{\`e}re, Th{\'e}ophane Weber, David Raposo, Adam Santoro, Laurent
  Orseau, Tom Eccles, et~al.
\newblock An investigation of model-free planning.
\newblock In {\em International Conference on Machine Learning}, pages
  2464--2473. PMLR, 2019.

\bibitem{DBLP:conf/iclr/HamrickFBGVWABV21}
Jessica~B. Hamrick, Abram~L. Friesen, Feryal Behbahani, Arthur Guez, Fabio
  Viola, Sims Witherspoon, Thomas Anthony, Lars~Holger Buesing, Petar
  Velickovic, and Theophane Weber.
\newblock On the role of planning in model-based deep reinforcement learning.
\newblock In {\em 9th International Conference on Learning Representations,
  {ICLR} 2021, Virtual Event, Austria, May 3-7, 2021}. OpenReview.net, 2021.

\bibitem{hassabis2017neuroscience}
Demis Hassabis, Dharshan Kumaran, Christopher Summerfield, and Matthew
  Botvinick.
\newblock Neuroscience-inspired artificial intelligence.
\newblock {\em Neuron}, 95(2):245--258, 2017.

\bibitem{hasselt2010double}
Hado Hasselt.
\newblock Double q-learning.
\newblock {\em Advances in neural information processing systems},
  23:2613--2621, 2010.

\bibitem{hollerman2000involvement}
Jeffrey~R Hollerman, Leon Tremblay, and Wolfram Schultz.
\newblock Involvement of basal ganglia and orbitofrontal cortex in
  goal-directed behavior.
\newblock {\em Progress in brain research}, 126:193--215, 2000.

\bibitem{DBLP:conf/iclr/JayaramanEEL19}
Dinesh Jayaraman, Frederik Ebert, Alexei~A. Efros, and Sergey Levine.
\newblock Time-agnostic prediction: Predicting predictable video frames.
\newblock In {\em 7th International Conference on Learning Representations,
  {ICLR} 2019, New Orleans, LA, USA, May 6-9, 2019}. OpenReview.net, 2019.

\bibitem{DBLP:conf/ijcai/Kaelbling93}
Leslie~Pack Kaelbling.
\newblock Learning to achieve goals.
\newblock In Ruzena Bajcsy, editor, {\em Proceedings of the 13th International
  Joint Conference on Artificial Intelligence. Chamb{\'{e}}ry, France, August
  28 - September 3, 1993}, pages 1094--1099. Morgan Kaufmann, 1993.

\bibitem{DBLP:conf/nips/KimAB19}
Taesup Kim, Sungjin Ahn, and Yoshua Bengio.
\newblock Variational temporal abstraction.
\newblock In Hanna~M. Wallach, Hugo Larochelle, Alina Beygelzimer, Florence
  d'Alch{\'{e}}{-}Buc, Emily~B. Fox, and Roman Garnett, editors, {\em Advances
  in Neural Information Processing Systems 32: Annual Conference on Neural
  Information Processing Systems 2019, NeurIPS 2019, December 8-14, 2019,
  Vancouver, BC, Canada}, pages 11566--11575, 2019.

\bibitem{DBLP:conf/nips/KurutachTYRA18}
Thanard Kurutach, Aviv Tamar, Ge~Yang, Stuart~J. Russell, and Pieter Abbeel.
\newblock Learning plannable representations with causal infogan.
\newblock In Samy Bengio, Hanna~M. Wallach, Hugo Larochelle, Kristen Grauman,
  Nicol{\`{o}} Cesa{-}Bianchi, and Roman Garnett, editors, {\em Advances in
  Neural Information Processing Systems 31: Annual Conference on Neural
  Information Processing Systems 2018, NeurIPS 2018, December 3-8, 2018,
  Montr{\'{e}}al, Canada}, pages 8747--8758, 2018.

\bibitem{lipovetzky2012width}
Nir Lipovetzky and Hector Geffner.
\newblock Width and serialization of classical planning problems.
\newblock In {\em ECAI 2012}, pages 540--545. IOS Press, 2012.

\bibitem{DBLP:conf/icml/LiuKTAT20}
Kara Liu, Thanard Kurutach, Christine Tung, Pieter Abbeel, and Aviv Tamar.
\newblock Hallucinative topological memory for zero-shot visual planning.
\newblock In {\em Proceedings of the 37th International Conference on Machine
  Learning, {ICML} 2020, 13-18 July 2020, Virtual Event}, volume 119 of {\em
  Proceedings of Machine Learning Research}, pages 6259--6270. {PMLR}, 2020.

\bibitem{DBLP:journals/corr/abs-2001-08210}
Yinhan Liu, Jiatao Gu, Naman Goyal, Xian Li, Sergey Edunov, Marjan
  Ghazvininejad, Mike Lewis, and Luke Zettlemoyer.
\newblock Multilingual denoising pre-training for neural machine translation.
\newblock {\em CoRR}, abs/2001.08210, 2020.

\bibitem{milos2019uncertainty}
Piotr Mi{\l}o{\'s}, {\L}ukasz Kuci{\'n}ski, Konrad Czechowski, Piotr
  Kozakowski, and Maciek Klimek.
\newblock Uncertainty-sensitive learning and planning with ensembles.
\newblock {\em arXiv preprint arXiv:1912.09996}, 2019.

\bibitem{nair2019hierarchical}
Suraj Nair and Chelsea Finn.
\newblock Hierarchical foresight: Self-supervised learning of long-horizon
  tasks via visual subgoal generation.
\newblock In {\em 8th International Conference on Learning Representations,
  {ICLR} 2020, Addis Ababa, Ethiopia, April 26-30, 2020}. OpenReview.net, 2020.

\bibitem{DBLP:conf/nips/NasirianyPLL19}
Soroush Nasiriany, Vitchyr Pong, Steven Lin, and Sergey Levine.
\newblock Planning with goal-conditioned policies.
\newblock In Hanna~M. Wallach, Hugo Larochelle, Alina Beygelzimer, Florence
  d'Alch{\'{e}}{-}Buc, Emily~B. Fox, and Roman Garnett, editors, {\em Advances
  in Neural Information Processing Systems 32: Annual Conference on Neural
  Information Processing Systems 2019, NeurIPS 2019, December 8-14, 2019,
  Vancouver, BC, Canada}, pages 14814--14825, 2019.

\bibitem{parascandolo2020divide}
Giambattista Parascandolo, Lars Buesing, Josh Merel, Leonard Hasenclever, John
  Aslanides, Jessica~B Hamrick, Nicolas Heess, Alexander Neitz, and Theophane
  Weber.
\newblock Divide-and-conquer monte carlo tree search for goal-directed
  planning.
\newblock {\em arXiv preprint arXiv:2004.11410}, 2020.

\bibitem{paul2019learning}
Sujoy Paul, Jeroen Vanbaar, and Amit Roy-Chowdhury.
\newblock Learning from trajectories via subgoal discovery.
\newblock {\em Advances in Neural Information Processing Systems},
  32:8411--8421, 2019.

\bibitem{DBLP:conf/nips/PertschREZJFL20}
Karl Pertsch, Oleh Rybkin, Frederik Ebert, Shenghao Zhou, Dinesh Jayaraman,
  Chelsea Finn, and Sergey Levine.
\newblock Long-horizon visual planning with goal-conditioned hierarchical
  predictors.
\newblock In Hugo Larochelle, Marc'Aurelio Ranzato, Raia Hadsell,
  Maria{-}Florina Balcan, and Hsuan{-}Tien Lin, editors, {\em Advances in
  Neural Information Processing Systems 33: Annual Conference on Neural
  Information Processing Systems 2020, NeurIPS 2020, December 6-12, 2020,
  virtual}, 2020.

\bibitem{DBLP:conf/l4dc/PertschRYZDDLJ20}
Karl Pertsch, Oleh Rybkin, Jingyun Yang, Shenghao Zhou, Konstantinos~G.
  Derpanis, Kostas Daniilidis, Joseph~J. Lim, and Andrew Jaegle.
\newblock Keyframing the future: Keyframe discovery for visual prediction and
  planning.
\newblock In Alexandre~M. Bayen, Ali Jadbabaie, George~J. Pappas, Pablo~A.
  Parrilo, Benjamin Recht, Claire~J. Tomlin, and Melanie~N. Zeilinger, editors,
  {\em Proceedings of the 2nd Annual Conference on Learning for Dynamics and
  Control, {L4DC} 2020, Online Event, Berkeley, CA, USA, 11-12 June 2020},
  volume 120 of {\em Proceedings of Machine Learning Research}, pages 969--979.
  {PMLR}, 2020.

\bibitem{DBLP:conf/icml/PitisCZSB20}
Silviu Pitis, Harris Chan, Stephen Zhao, Bradly~C. Stadie, and Jimmy Ba.
\newblock Maximum entropy gain exploration for long horizon multi-goal
  reinforcement learning.
\newblock In {\em Proceedings of the 37th International Conference on Machine
  Learning, {ICML} 2020, 13-18 July 2020, Virtual Event}, volume 119 of {\em
  Proceedings of Machine Learning Research}, pages 7750--7761. {PMLR}, 2020.

\bibitem{polu2020generative}
Stanislas Polu and Ilya Sutskever.
\newblock Generative language modeling for automated theorem proving.
\newblock {\em arXiv preprint arXiv:2009.03393}, 2020.

\bibitem{RacaniereWRBGRB17}
S{\'{e}}bastien Racani{\`{e}}re, Theophane Weber, David~P. Reichert, Lars
  Buesing, Arthur Guez, Danilo~Jimenez Rezende, Adri{\`{a}}~Puigdom{\`{e}}nech
  Badia, Oriol Vinyals, Nicolas Heess, Yujia Li, Razvan Pascanu, Peter
  Battaglia, Demis Hassabis, David Silver, and Daan Wierstra.
\newblock Imagination-augmented agents for deep reinforcement learning.
\newblock In {\em NIPS}, 2017.

\bibitem{rokicki2014god}
Tomas Rokicki and M~Davidson.
\newblock God’s number is 26 in the quarter-turn metric, 2014.

\bibitem{russell2002artificial}
Stuart Russell and Peter Norvig.
\newblock Artificial intelligence: A modern approach. ed. 3.
\newblock 2010.

\bibitem{SavinovDK18}
Nikolay Savinov, Alexey Dosovitskiy, and Vladlen Koltun.
\newblock Semi-parametric topological memory for navigation.
\newblock In {\em 6th International Conference on Learning Representations,
  {ICLR} 2018, Vancouver, BC, Canada, April 30 - May 3, 2018, Conference Track
  Proceedings}. OpenReview.net, 2018.

\bibitem{silver2016mastering}
David Silver, Aja Huang, Chris~J Maddison, Arthur Guez, Laurent Sifre, George
  Van Den~Driessche, Julian Schrittwieser, Ioannis Antonoglou, Veda
  Panneershelvam, Marc Lanctot, et~al.
\newblock Mastering the game of go with deep neural networks and tree search.
\newblock {\em nature}, 529(7587):484--489, 2016.

\bibitem{Silver2018}
David Silver, Thomas Hubert, Julian Schrittwieser, Ioannis Antonoglou, Thore
  Graepel, Timothy Lillicrap, Karen Simonyan, and Demis Hassabis.
\newblock {A general reinforcement learning algorithm that masters chess,
  shogi, and Go through self-play}.
\newblock {\em Science}, 1144:1140--1144, 2018.

\bibitem{Silver2017MasteringCA}
David Silver, Thomas Hubert, Julian Schrittwieser, Ioannis Antonoglou, Matthew
  Lai, Arthur Guez, Marc Lanctot, Laurent Sifre, Dharshan Kumaran, Thore
  Graepel, Timothy~P. Lillicrap, Karen Simonyan, and Demis Hassabis.
\newblock Mastering chess and shogi by self-play with a general reinforcement
  learning algorithm.
\newblock {\em ArXiv}, abs/1712.01815, 2017.

\bibitem{Silver2017}
David Silver, Julian Schrittwieser, Karen Simonyan, Ioannis Antonoglou, Aja
  Huang, Arthur Guez, Thomas Hubert, Lucas Baker, Matthew Lai, Adrian Bolton,
  Yutian Chen, Timothy Lillicrap, Fan Hui, Laurent Sifre, George {Van Den
  Driessche}, Thore Graepel, and Demis Hassabis.
\newblock {Mastering the game of Go without human knowledge}.
\newblock {\em Nature}, 2017.

\bibitem{DBLP:conf/corl/SteinBR18}
Gregory~J. Stein, Christopher Bradley, and Nicholas Roy.
\newblock Learning over subgoals for efficient navigation of structured,
  unknown environments.
\newblock In {\em 2nd Annual Conference on Robot Learning, CoRL 2018,
  Z{\"{u}}rich, Switzerland, 29-31 October 2018, Proceedings}, volume~87 of
  {\em Proceedings of Machine Learning Research}, pages 213--222. {PMLR}, 2018.

\bibitem{DBLP:conf/atal/SuttonMDDPWP11}
Richard~S. Sutton, Joseph Modayil, Michael Delp, Thomas Degris, Patrick~M.
  Pilarski, Adam White, and Doina Precup.
\newblock Horde: a scalable real-time architecture for learning knowledge from
  unsupervised sensorimotor interaction.
\newblock In Liz Sonenberg, Peter Stone, Kagan Tumer, and Pinar Yolum, editors,
  {\em 10th International Conference on Autonomous Agents and Multiagent
  Systems {(AAMAS} 2011), Taipei, Taiwan, May 2-6, 2011, Volume 1-3}, pages
  761--768. {IFAAMAS}, 2011.

\bibitem{DBLP:journals/ai/SuttonPS99}
Richard~S. Sutton, Doina Precup, and Satinder~P. Singh.
\newblock Between mdps and semi-mdps: {A} framework for temporal abstraction in
  reinforcement learning.
\newblock {\em Artif. Intell.}, 112(1-2):181--211, 1999.

\bibitem{vaswani2017attention}
Ashish Vaswani, Noam Shazeer, Niki Parmar, Jakob Uszkoreit, Llion Jones,
  Aidan~N. Gomez, Lukasz Kaiser, and Illia Polosukhin.
\newblock Attention is all you need.
\newblock In Isabelle Guyon, Ulrike von Luxburg, Samy Bengio, Hanna~M. Wallach,
  Rob Fergus, S.~V.~N. Vishwanathan, and Roman Garnett, editors, {\em Advances
  in Neural Information Processing Systems 30: Annual Conference on Neural
  Information Processing Systems 2017, December 4-9, 2017, Long Beach, CA,
  {USA}}, pages 5998--6008, 2017.

\bibitem{DBLP:conf/nips/VaswaniSPUJGKP17}
Ashish Vaswani, Noam Shazeer, Niki Parmar, Jakob Uszkoreit, Llion Jones,
  Aidan~N. Gomez, Lukasz Kaiser, and Illia Polosukhin.
\newblock Attention is all you need.
\newblock In Isabelle Guyon, Ulrike von Luxburg, Samy Bengio, Hanna~M. Wallach,
  Rob Fergus, S.~V.~N. Vishwanathan, and Roman Garnett, editors, {\em Advances
  in Neural Information Processing Systems 30: Annual Conference on Neural
  Information Processing Systems 2017, December 4-9, 2017, Long Beach, CA,
  {USA}}, pages 5998--6008, 2017.

\bibitem{Vezhnevets2017FeUdalNF}
Alexander~Sasha Vezhnevets, Simon Osindero, Tom Schaul, Nicolas Manfred~Otto
  Heess, Max Jaderberg, David Silver, and Koray Kavukcuoglu.
\newblock Feudal networks for hierarchical reinforcement learning.
\newblock {\em ArXiv}, abs/1703.01161, 2017.

\bibitem{Vinyals2019}
Oriol Vinyals, Igor Babuschkin, Wojciech~M. Czarnecki, Micha{\"e}l Mathieu,
  Andrew Dudzik, Junyoung Chung, David~H. Choi, Richard Powell, Timo Ewalds,
  Petko Georgiev, Junhyuk Oh, Dan Horgan, Manuel Kroiss, Ivo Danihelka, Aja
  Huang, Laurent Sifre, Trevor Cai, John~P. Agapiou, Max Jaderberg,
  Alexander~S. Vezhnevets, R{\'e}mi Leblond, Tobias Pohlen, Valentin Dalibard,
  David Budden, Yury Sulsky, James Molloy, Tom~L. Paine, Caglar Gulcehre, Ziyu
  Wang, Tobias Pfaff, Yuhuai Wu, Roman Ring, Dani Yogatama, Dario W{\"u}nsch,
  Katrina McKinney, Oliver Smith, Tom Schaul, Timothy Lillicrap, Koray
  Kavukcuoglu, Demis Hassabis, Chris Apps, and David Silver.
\newblock Grandmaster level in starcraft ii using multi-agent reinforcement
  learning.
\newblock {\em Nature}, 575(7782):350--354, Nov 2019.

\bibitem{DBLP:journals/ai/Wilkins80}
David Wilkins.
\newblock Using patterns and plans in chess.
\newblock {\em Artif. Intell.}, 14(2):165--203, 1980.

\bibitem{DBLP:journals/corr/abs-1910-03771}
Thomas Wolf, Lysandre Debut, Victor Sanh, Julien Chaumond, Clement Delangue,
  Anthony Moi, Pierric Cistac, Tim Rault, R{\'{e}}mi Louf, Morgan Funtowicz,
  and Jamie Brew.
\newblock Huggingface's transformers: State-of-the-art natural language
  processing.
\newblock {\em CoRR}, abs/1910.03771, 2019.

\bibitem{wu2020int}
Yuhuai Wu, Albert Jiang, Jimmy Ba, and Roger Grosse.
\newblock Int: An inequality benchmark for evaluating generalization in theorem
  proving.
\newblock {\em arXiv preprint arXiv:2007.02924}, 2020.

\bibitem{DBLP:journals/corr/abs-2011-12491}
Lunjun Zhang, Ge~Yang, and Bradly~C. Stadie.
\newblock World model as a graph: Learning latent landmarks for planning.
\newblock {\em CoRR}, abs/2011.12491, 2020.

\end{thebibliography}
\bibliographystyle{plain}

\newpage

\appendix

\section{MCTS}\label{sec:mcts_appendix}

\subsection{MCTS-\abbrv{} algorithm}\label{sec:mcts_algorithm_appendix}

In Algorithm \ref{alg:generic_mcts} we present a general MCTS solver based on AlphaZero. Solver repeatedly queries the planner for a list of actions and executes them one by one. Baseline planner returns only a single action at a time, whereas MCTS-\abbrv{} gives around $k$ actions -- to reach the desired subgoal (number of actions depends on a subgoal distance, which not always equals $k$ in practice).

MCTS-\abbrv{} operates on a high-level subgoal graph: nodes are subgoals proposed by the generator (see Algorithm \ref{alg:subgoal_generator}) and edges -- lists of actions informing how to move from one subgoal to another (computed by the low-level conditional policy in Algorithm \ref{alg:conditional_policy}). The graph structure is represented by $tree$ variable. For every subgoal, it keeps up to $C_3$ best nearby subgoals (according to generator scores) along with a mentioned list of actions and sum of rewards to obtain while moving from the parent to the child subgoal.

Most of MCTS implementation is shared between MCTS-\abbrv{} and AlphaZero baseline, as we can treat the behavioral-cloning policy as a subgoal generator with $k = 1$. All the differences between MCTS-\abbrv{} and the baseline are encapsulated in $\Call{gen\_children}{}$ function (Algorithms \ref{alg:gen_children_mcts_ksubs} and \ref{alg:gen_children_alpha_zero}). To generate children subgoals MCTS-\abbrv{} runs subgoal generator and low-level conditional policy, whereas the baseline uses behavioral cloning policy for that purpose.

\begin{algorithm}[H]
    \caption{MCTS solver (common for AlphaZero baseline and MCTS-\abbrv{})}
    \label{alg:generic_mcts}
\begin{minipage}[h]{.46\textwidth}
\begin{tabular}{ l c l }
    \textbf{Require: }
    & $L_a$ & action limit \\
    & $L_p$ & planner calls limit \\
    & $P$ & planning passes \\
    & $\gamma$ & discount factor  \\
    & $c_{puct}$ & exploration weight \\
    & $\tau$ & sampling temperature \\
    & $V$ & value function \\
    & $env$ & environment \\
    & $M$ & environment model \\
	\textbf{Use:}
	& $tree$ & tree structure \\
	& $N(s, i)$ & visit count \\
	& $W(s, i)$ & total child-value \\
	& $Q(s, i)$ & mean child-value \\
	& $\pi_e$ & exploration policy \\
\end{tabular}
\begin{algorithmic}
    \State $\texttt{\# Initialize $N, W, Q$ to zero}$
    \Function{solver}{}
		\State $s \gets env.\Call{reset}{ }$
		\State $\mathtt{solution} \gets \mathtt{[\,]}$  \Comment{List of actions}
		\For{$1\ldots L_p$}
			\State $\mathtt{actions} \gets$ \Call{planner}{s}
			\For{$a \textbf{ in } \mathtt{actions}$}
			    \State $s', r \gets env.\Call{step}{a}$
			    \State $\mathtt{solution}.\Call{append}{a}$
			    \State $s \gets s'$
		    \EndFor
		    \If{$\mathtt{solution}.\Call{length}{ } > L_a$} 
		        \State \textbf{return} $\mathtt{None}$
	        \EndIf
	        \If{$env.\Call{solved}{ }$}
	            \State \textbf{return} $\mathtt{solution}$
            \EndIf
		\EndFor
		\State \Return $\mathtt{None}$
    \EndFunction
\end{algorithmic}
\begin{algorithmic}
    \Function{planner}{$\mathtt{state}$}
        \For{$1\ldots P$}  
            \State $\mathtt{path,~leaf} \gets$ \Call{select}{$\mathtt{state}$}
            \State \Call{expand}{$\mathtt{leaf}$} 
            \State \Call{update}{$\mathtt{path,~leaf}$}
        \EndFor
        \State \Return \Call{choose\_actions}{$\mathtt{state}$}
    \EndFunction
\end{algorithmic}
\end{minipage}
\begin{minipage}[h]{0.54\textwidth}
\begin{algorithmic}
    \Function{select}{$\mathtt{state}$}
        \State $s \gets \mathtt{state}$
        \State $\mathtt{path} \gets [\,]$
        \While{$s~\text{belongs to}~tree$} 
            \State $i \gets$ \Call{select\_child}{$s$} 
            \State $s', r, \mathtt{actions} \gets tree[s][i]$
            \State $\mathtt{path}.\Call{append}{(s, i, r)}$
            \State $s \gets s'$
        \EndWhile
        \State \Return $\mathtt{path},~ s$
    \EndFunction
\end{algorithmic}
\begin{algorithmic}
    \Function{expand}{$\mathtt{leaf}$}
        \State $\mathtt{children}, \mathtt{probs} \gets$ \Call{gen\_children}{$\mathtt{leaf}$}
        \State $tree[\mathtt{leaf}] \gets \mathtt{children}$
        \State $\pi(\mathtt{leaf}, \cdot) \gets \mathtt{probs}$
        \For{$i \gets 1$ to $\mathtt{children}.\Call{length}{ }$}
            \State $s', r, \mathtt{actions} \gets tree[\mathtt{leaf}][i]$
            \State $W(\mathtt{leaf}, i) \gets r + \gamma * V(s')$
            \State $N(\mathtt{leaf}, i) \gets 1$
            \State $Q(\mathtt{leaf}, i) \gets W(\mathtt{leaf}, i)$
        \EndFor
    \EndFunction
\end{algorithmic}
\begin{algorithmic}
    \Function{update}{$\mathtt{path,~leaf}$}
        \State $\mathtt{quality} \gets V(\mathtt{leaf})$ 
        \For{$s, i, r \gets \mathtt{reversed(path)}$}
            \State $\mathtt{quality} \gets r + \gamma * \mathtt{quality}$
            \State $W(s, i) \gets W(s, i) + \mathtt{quality}$
            \State $N(s, i) \gets N(s, i) + 1$
            \State $Q(s, i) \gets \frac{W(s, i)}{N(s, i)}$
        \EndFor
    \EndFunction
\end{algorithmic}
\begin{algorithmic}
    \Function{select\_child}{$s$}
        \State $U(s, i) \gets \sqrt{\sum_{i'}N(s,i')}/(1 + N(s,i))$
        \State $i \gets argmax_i (Q(s,i) + c_{puct} \pi_e(s, i) U(s, i))$
        \State \Return $i$
    \EndFunction
\end{algorithmic}
\begin{algorithmic}
    \Function{choose\_actions}{$s$}
        \State $i \sim softmax \big( \frac{1}{\tau} \log N(s, \cdot) \big)$
        \State $s', r, \mathtt{actions} \gets tree[s][i]$
        \State \Return $\mathtt{actions}$
    \EndFunction
\end{algorithmic}
\end{minipage}
\end{algorithm}

\begin{minipage}[t]{0.5\textwidth}
\begin{algorithm}[H]
    \caption{GEN\_CHILDREN for MCTS-\abbrv{}}
    \label{alg:gen_children_mcts_ksubs}
\textit{For functions GET\_PATH and SUB\_GENERATE see Algorithms \ref{alg:conditional_policy} and \ref{alg:subgoal_generator}.}
\begin{algorithmic}\small
    \Function{gen\_children}{$\mathtt{state}$}
        \State $s \gets \mathtt{state}$
        \State $\mathtt{children} \gets [\,]$
        \State $\mathtt{probs} \gets [\,]$
        \For{$\mathtt{subgoal}, \mathtt{prob} \gets$ \Call{sub\_generate}{$s$}}
            \State $\mathtt{actions} \gets $ \Call{get\_path}{$s, \mathtt{subgoal}$}
            \If{$\mathtt{actions}.$\Call{empty}{ }}{ $\mathtt{continue}$}\EndIf
            \State $r \gets M.$\Call{reward\_sum}{$s, \mathtt{actions}$}
            \State $\mathtt{children}$.\Call{append}{$(\mathtt{subgoal}, r, \mathtt{actions})$}
            \State $\mathtt{probs}.$\Call{append}{$\mathtt{prob}$}
        \EndFor
        \State \Return $\mathtt{children,~probs}$
    \EndFunction
\end{algorithmic}
\end{algorithm}
\end{minipage}
\begin{minipage}[t]{0.5\textwidth}
\begin{algorithm}[H]
    \caption{GEN\_CHILDREN for AlphaZero}
    \label{alg:gen_children_alpha_zero}
\begin{tabular}{ l c l }
    \textbf{Require: }
    & $\pi_b$ & behavioral cloning policy \\
\end{tabular}
\begin{algorithmic}\small
    \Function{gen\_children}{$\mathtt{state}$}
        \State $s \gets \mathtt{state}$
        \State $\mathtt{children} \gets [\,]$
        \State $\mathtt{probs} \gets [\,]$
        \For{$a, \mathtt{prob} \gets \pi_b.$\Call{gen\_actions}{$s$}}
            \State $s', r \gets M.$\Call{next\_state\_reward}{$s, a$}
            \State $\mathtt{children}$.\Call{append}{$(s', r, [a])$}
            \State $\mathtt{probs}.$\Call{append}{$\mathtt{prob}$}
        \EndFor
        \State \Return $\mathtt{children,~probs}$
    \EndFunction
\end{algorithmic}
\end{algorithm}
\end{minipage}

Variables $tree, N, W, Q, \pi_e$ are reused across subsequent planner invocations within a single solver run. We limit the number of planner calls $L_p$ for better control over the computational budget for MCTS-\abbrv{}. For MCTS pseudocode we assume a slightly modified version of $\Call{sub\_generate}{}$ function (defined originally in Algorithm \ref{alg:subgoal_generator}). We presume that the function along with subgoals returns also their respective probabilities -- as MCTS needs them to guide exploration.

\subsection{Detailed results of MCTS-\abbrv{}}\label{sec:detailed_results_mcts_appendix}

We evaluate MCTS-based approaches on INT proofs of length 15. We set $c_{puct} = \tau = 1$ and $\gamma = 0.99$. We tuned $P$ on MCTS (AlphaZero) baseline and we run three variants with $P \in \{5, 15, 50\}$. We run MCTS-\abbrv{} ($k = 3$) with the same set of parameters and with  \abbrv{}-specific parameters fixed to $C_2 = C_3 = 4$ (in order to match the setup for corresponding INT BF-\abbrv{} experiments).

We limit the maximum number of actions to $L_a = 24$ for both methods. 
Having the same number of planning passes $P$, during a single call MCTS-\abbrv{} visits $k$-times more new states than the baseline (because of states visited by the low-level conditional policy). Therefore, to ensure a similar computational budget, we limit the number of planner calls to $L_p = 8$ for MCTS-\abbrv{} and to $L_p = 24$ for the baseline -- so the number of states visited over the course of a single solver run is similar for both methods. 

Top-left part of Figure \ref{fig:main_results} illustrates results of MCTS experiments. For every number of planning passes~$P$, MCTS-\abbrv{} has significantly higher success rate than the corresponding baseline experiment. The highest difference is $0.52$ for $P = 5$ ($0.88$ for MCTS-\abbrv{}, $0.36$ for the baseline) and slightly decreases with increasing number of passes to still impressive $0.43$ for $P = 50$ ($0.91$ for MCTS-\abbrv{}, $0.48$ for the baseline). Comparing MCTS-\abbrv{} for $P = 5$ with the baseline for $P = 50$, shows advantage of our method still by a significant margin of $0.40$, despite having 10 times smaller computational budget.

MCTS-\abbrv{} performed better also in terms of run time. For every tested $P$ it was at least twice as fast as the corresponding baseline experiment.

High effectiveness of MCTS-\abbrv{}, in terms of both search success rate as well as run time, shows that our \abbrv{} method is not specific to BestFS planner, but potentially allows to boost a wide range of other planners.



\section{Architectures and hyperparameters}\label{sec:architectures_and_hyperparameters}
\subsection{Transformer} \label{sec:transformer_architecture}




For INT and Rubik we use mBART \cite{DBLP:journals/corr/abs-2001-08210} -- one of the state-of-the-art sequence-to-sequence transformer architectures. To speed up training and inference we use its lightweight version. We reduced the dimensionality of the model, so the number of learned parameters decreased from the original 680M to 45M. The set of our hyperparameters matches the values proposed in \cite{vaswani2017attention}: we used 6 layers of encoder and 6 layers of decoder; we adjusted model's dimension to 512 and number of attention heads to 8; the inner-layer of position-wise fully connected networks had dimensionality 2048. The difference in our model's size compared to 65M parameters reported in \cite{vaswani2017attention} results from vocabulary size. For our problems, it is enough to have 10-70 distinct tokens, whereas natural language models require a much larger vocabulary (tens of thousands of tokens).

For inference we used number of beams equal to 16 on INT and 32 on Rubik's Cube.

\subsection{Sokoban} \label{sec:appendix_sokoban_generator}

In Sokoban, we use three neural network architectures: for generating subgoals,  for assigning value and one for baseline policy.

We took the value function network architecture from \cite{milos2019uncertainty}. 
For the subgoal generator network we used the same convolutional architecture as in \cite{milos2019uncertainty}, with two exceptions. First, instead of predicting single regression target we predicted distribution over $d \times d \times 7 + 1$ classes. Secondly, we added batch norm layers between convolutional layers to speed up training. To make the comparison between BestFS and BF-\abbrv{} fair, we also evaluated the training of expert from \cite{milos2019uncertainty} with additional batch norm layers, but it turned out to actually hurt the performance of the expert. The architecture for baseline policy was the same as in \cite{milos2019uncertainty} with only one modification: it predicts one of our actions instead of a single number.




\section{Data processing and datasets}\label{sec:data_processing_appendix}


\subsection{Sokoban}\label{sec:appendix_datasets_sokoban}

\textbf{Dataset}. We collected expert datasets using an RL agent (MCTS-based) from \cite{milos2019uncertainty}. Precisely, we trained $3$ agents on Sokoban boards of different sizes ($12\times12$, $16\times16$ and $20\times20$, all with four boxes). During the training process, we collected all successful trajectories, in the form of sequences of consecutive states. The number of trajectories in our datasets were: $154 000$ for $12\times12$ boards, $45 000$ for $16\times 16$ boards and $21 500$ for $20\times 20$ boards. The difference comes from the fact that 
the larger boards take more time to solve, hence fewer data is collected in the same time span. 

\begin{figure}[t]
\includegraphics[width=0.90\textwidth]{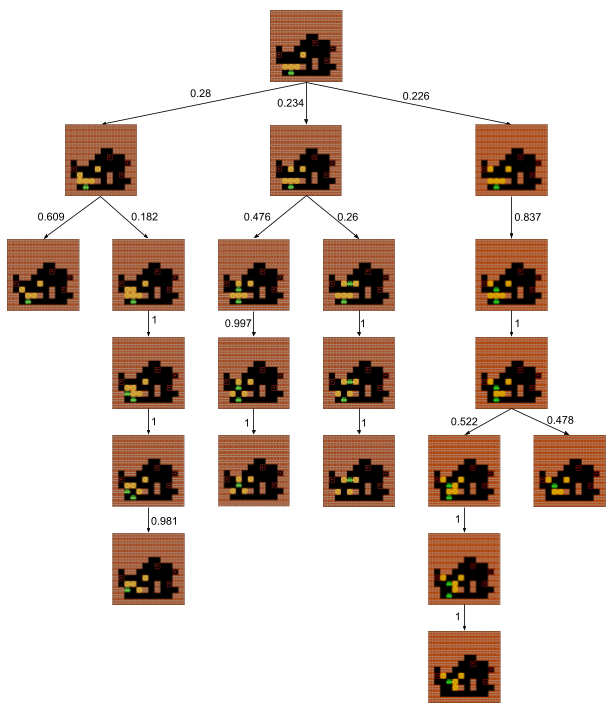}
\caption{A detailed view of subgoal generation for Sokoban. Arrow represent probabilities of a given modification. Final subgoals are located in the leaves.}
\label{fig:sokoban_generation}
\end{figure}

\textbf{Subgoal  generation}. For a given $\mathtt{state}$ the generation of subgoals is depicted in Algorithm \ref{alg:sokoban_subgoal_generator}. We maintain a queue of modified states (MS). Iteratively we take a MS from queue, concatenate it with $\mathtt{state}$ and pass through subgoal generator network ($\mathtt{subgoal\_net.\Call{sorted\_predictions}{}}$). This produces a probability distribution over candidates for further modifications of given MS. We take the most probable candidates, apply each of them to MS, and add the new modified states to the queue. If among the best subgoal generator network predictions there is a special "valid subgoal" token (encoded with $d \times d \times 7 + 1$), we put MS to subgoal candidates list ($\mathtt{subgoals\_and\_probs}$). During this process, each MS is assigned the probability, which is a product of probabilities of modifications, leading to this MS. When the queue is empty, we take subgoal candidates and choose the ones with the highest probability such that the target probability ($C_4$) is reached (similar to Algorithm \ref{alg:subgoal_generator}). The generation of subgoals for a given state is illustrated in Figure \ref{fig:sokoban_generation}.

This process is designed to be computationally efficient. The majority of subgoals differ from the input by only several pixels, which leads to short paths of point-wise modifications. Note, that we do not utilize any Sokoban-specific assumptions. 

\textbf{Datapoints for training}. Preparing data points for the training of the generator is described in Algorithm \ref{alg:sokoban_targets}. For each trajectory in the dataset, we choose randomly 10\% of state pairs for the training (we do not use all states from a trajectory in order to reduce the correlation in the data). 

\textbf{Low-level conditional policy}. In Algorithm \ref{alg:sokoban_conditional_policy}
we describe the BFS-based algorithm that verifies subgoals in Sokoban. 

\textbf{Performance of RL agent}. We observed that each of the three RL agents we used (for $12\times12$, $16\times16$ and $20\times20$ boards), had a significantly lower success rate than our method's counterparts (that learns from these agents). For $12\times12$ boards it could solve around 78\% of problems, for $16\times16$ boards it dropped to 67\% and for $20\times 20$ it was only 60\%.


\begin{minipage}[t]{1.\textwidth}
\begin{algorithm}[H]
    \caption{Sokoban subgoal generator}
    \label{alg:sokoban_subgoal_generator}
\begin{tabular}{ l c l }
    \textbf{Require: }
    & $\mathtt{d}$ & dimension of a board \\
    & $\mathtt{internal\_cl}$ & a number between 0 and 1 \\
    & $\mathtt{subgoal\_net}$ & CNN returning distribution over modifications. \\
\end{tabular}
\begin{algorithmic}
    \Function{generate\_subgoals}{$\mathtt{state}$}
    \State $\mathtt{subgoals\_and\_probs} \gets []$
    \State $\mathtt{q \gets Queue()}$ \Comment{FIFO queue}
    \State $\mathtt{q.\Call{insert}{(state, 1)}}$
    
    \While{not $\mathtt{q}$ not empty}
        \State $\mathtt{modified\_state, parent\_prob \gets q.\Call{pop}{~}}$
        \State $\mathtt{network\_input \gets \Call{concatenate}{state, modified\_state}}$
        \State $\mathtt{predictions, probs \gets subgoal\_net.\Call{sorted\_predictions}{network\_input}}$ \label{lst:line:network_prediction}

        \State $\mathtt{total\_p} \gets 0$
        \For{$\mathtt{prediction}, \mathtt{p} \in (\mathtt{predictions}, \mathtt{probs})$}
            \State \textbf{if} {$\mathtt{total\_p} \geq \mathtt{internal\_cl}$} \textbf{ then break}
            \State $\mathtt{total\_p} \gets \mathtt{total\_p} + \mathtt{p}$
            \If{$\mathtt{prediction = d \times d \times 7 + 1}$}
                \State $\mathtt{subgoals\_and\_probs}.\Call{add}{(\mathtt{modified\_state, parent\_prob \times p)}}$
            \Else
                \State $\mathtt{new\_modified\_state \gets \Call{Apply\_change}{modified\_state, prediction}}$
                \State $\mathtt{q.\Call{insert}{(new\_modified\_state, parent\_prob \times p)}}$
            \EndIf
        \EndFor
        \State $\mathtt{subgoals \gets \Call{sort\_by\_probability}{subgoals}}$
        
    \EndWhile
    \State $\mathtt{total\_p} \gets 0$
    \For{$\mathtt{subgoal}, \mathtt{p} \in \mathtt{subgoals\_and\_probs}$}
            \State \textbf{if} {$\mathtt{total\_p} > C_4$} \textbf{ then break}
            \State $\mathtt{subgoals}.\Call{add}{\mathtt{state}}$
            \State $\mathtt{total\_p} \gets \mathtt{total\_p} + \mathtt{p}$
        \EndFor
    
    \State \Return $\mathtt{subgoals}$
    \EndFunction
    
    \Function{Apply\_change}{$\mathtt{state, modification}$}
        \State \(\triangleright\)  $\mathtt{modification}$ is an integer in range $\mathtt{[1, d \times d \times 7]}$ encoding which pixel of $\mathtt{state}$ 
        \State \(\triangleright\) to change (and to which value).
        \State $\mathtt{row \gets \frac{modification}{d \times 7}}$ \Comment{Integer division}
        \State $\mathtt{column \gets \frac{modification - row \times d \times 7}{7}}$ \Comment{Integer division}
        \State $\mathtt{depth \gets modification - row \times d \times 7 - column \times 7}$
        \State $\mathtt{modified\_state \gets state}$
        \State $\mathtt{\Call{set\_to\_zeroes}{modified\_state[row, column]}} $
        \State $\mathtt{modified\_state[row, column, depth] \gets 1} $
        \State \Return $\mathtt{modified\_state}$
    \EndFunction
\end{algorithmic}
\end{algorithm}
\end{minipage}

\begin{minipage}[t]{1.\textwidth}
\begin{algorithm}[H]
    \caption{Sokoban generate network inputs and targets}
    \label{alg:sokoban_targets}
\begin{tabular}{ l c l }
    \textbf{Require: }
    & $\mathtt{d}$ & dimension of a board \\
\end{tabular}
\begin{algorithmic}
    \Function{generate\_inputs\_and\_targets}{$\mathtt{state}, \mathtt{subgoal}$}
    
    \State $\mathtt{inputs} \gets []$ \Comment{empty list}
    \State $\mathtt{targets} \gets []$ \Comment{empty list}
    
    \State $\mathtt{modified\_state} \gets \mathtt{state}$
    \State $\mathtt{input} \gets \Call{concatenate}{\mathtt{state}, \mathtt{modified\_state}}$
    \State $\mathtt{inputs}.\Call{append}{\mathtt{input}}$
    \State $\mathtt{target\_class\_num} \gets 0$
    \For{$\mathtt{i} \in 1 \dots \mathtt{d}$}
    \For{$\mathtt{j} \in 1 \dots \mathtt{d}$}
    \For{$\mathtt{c} \in 1 \dots 7$}
    \State $\mathtt{target\_class\_num} \gets \mathtt{target\_class\_num} + 1$
    \If{$\mathtt{subgoal[i, j, c]} = 1~ \Call{and}{}~ 
    \mathtt{modified\_state[i, j, c]} = 0$}
            \State $\mathtt{targets}.\Call{append}{\mathtt{target\_class\_num}}$
            \State \(\triangleright\) Numpy notation, replace pixel values on position $\mathtt{(i,j)}$  with values 
            \State \(\triangleright\)  from $\mathtt{subgoal}$
            \State $\mathtt{modified\_state[i, j, :]} \gets \mathtt{subgoal[i, j, :]}$ 
            \State $\mathtt{input} \gets \Call{concatenate}{\mathtt{state}, \mathtt{modified\_state}}$
            \State $\mathtt{inputs}.\Call{append}{\mathtt{input}}$
    \EndIf
    \EndFor
    \EndFor
    \EndFor
    \State \(\triangleright\)  Last target: no more changes to the $\mathtt{modified\_state}$ are needed (class enumerated 
    \State \(\triangleright\) with $\mathtt{d \times d \times 7 + 1}$) 
    \State $\mathtt{targets}.\Call{append}{\mathtt{d \times d \times 7 + 1}}$
    \State \Return $\mathtt{inputs, targets}$
    \EndFunction
\end{algorithmic}
\end{algorithm}
\end{minipage}

\begin{algorithm}[H]
    \caption{BFS low-level conditional policy}
    \label{alg:sokoban_conditional_policy}
\begin{tabular}{ l c l }
    \textbf{Require: }
    & $k$ & limit of steps \\
    & $M$ & model of the Sokoban environment \\
	\textbf{Use:}
	& $\mathtt{bfs\_queue}$ & BFS  queue; \\
	& & stores pairs of a state and the action path to it (from the root).
\end{tabular}
\begin{algorithmic}
    \State $\texttt{\# Initialize $\mathtt{bfs\_queue}$ to empty}$ 
    \Function{get\_path}{$\mathtt{s_0}$, $\mathtt{subgoal}$}
        \State $\mathtt{step} \gets 0$
        \State $\mathtt{bfs\_queue}.\Call{add}{(\mathtt{s_0}, [])}$
        \While{$\mathtt{bfs\_queue}$ not empty}
            \State $\mathtt{s}, \mathtt{action\_path}  \gets \mathtt{bfs\_queue}.\Call{pop}{~}$
            \For{$\mathtt{action} \in  \mathtt{action\_space}$}
                \State $\mathtt{s} \gets M.\Call{next\_state}{\mathtt{s, action}}$
                \State $\mathtt{action\_path}.\Call{append}{\mathtt{action}}$
                \If{$\mathtt{s} = \mathtt{subgoal}$}
                    \State \Return $\mathtt{action\_path}$
                \EndIf
                \If{$\Call{len}{\mathtt{action\_path}} < k$}
                    \State $\mathtt{bfs\_queue}.\Call{add}{(\mathtt{s}, \mathtt{action\_path})}$
                \EndIf
            \EndFor
        \EndWhile
        \State \Return $[]$
    \EndFunction
\end{algorithmic}
\end{algorithm}

\subsection{INT}
\textbf{State representation.} A state in INT consists of objectives to prove and ground truth assumptions, which are logic statements that are assumed to hold and can be used in proving. Each objective, as well as each ground truth assumption, is a mathematical statement. In our setup, as in the original paper \cite{wu2020int}, there is always only one objective to prove, but there may be a varying number of ground truth statements. 

Each mathematical statement can be converted to a string by a method $\mathtt{logic\_statement\_to\_seq\_string}$ in INT library. In our code, we represent the full state as a string in the following form (all symbols are tokens, not logical operations):
$$\#[\text{objective}]\&[\text{1st ground truth}]\&[\text{2nd ground truth}]\& \dots\ \&[k-\text{th ground truth}]\$ $$

For example, the state representation could look like this: 
$$\#(((b+b)*((b+b)*(b+b)))*((((b+b)+f)*(b+b))*(b+b))) \geq 0\&(b+f)=b\&(b+f)\geq 0\$ $$

\textbf{Action representation.} 
An action in INT consists of a chosen axiom (one from the set of ordered field axioms, see \cite[Appendix C]{wu2020int}) and a sequence of entities onto which the axiom will be applied. An entity is an algebraic expression that is a part of the larger statement. For example, $(a+b)$ is an entity, which is a part of $(a+b)\cdot c = (1+f)$. In our code, we represent entities by indicating the symbol of their mathematical operation, or if the entity is atomic (a single variable), by indicating the variable itself. More precisely, directly after the symbol of operation, we add a special character '$\sim$'. For example, we indicate $(a+b)$ inside $(a+b)\cdot c$ in the following way: $(a+\sim b)\cdot c=(1+f)$. Typically, in a logic statement there may be several entities that have the same text form, but are located in different positions, for example $(a+b)$ appears twice in $(a+b)\cdot (a+b) = (1+0)$. Our way of encoding actions unambiguously identifies every entity. If the action has more than one input, we use more different indicators.

\textbf{Low-level conditional policy input representation.} Low-level conditional policy takes as an input two states: $s$ and $s'$, where $s$ is the initial state and $s'$ is the subgoal. The input is constructed in the following way: first, we represent both $s$ and $s'$ as strings and then we find the difference (character delta) between these strings using the function ndiff from difflib\footnote{https://docs.python.org/3/library/difflib.html} Python library.  We observed that using the character delta, instead of the concatenation of $s$ and $s'$, significantly improved the performance.

\subsection{Rubik's Cube}

\textbf{State representation}
The state of the Rubik's Cube is determined by the arrangement of $54$ colored labels on its faces.
Therefore, to represent the observations we simply put the labels in a fixed order.
An example state is as follows:
$$?byywygrygobbrboorgwbowryoogywbggywrrroyrogyowwbrwwbbgg\$,$$
where the tokens \textit{b, g, o, r, w, y} stand for \textit{blue, green, orange, red, white}, and \textit{yellow}.
The consecutive blocks of 9 tokens correspond to consecutive faces of the cube.
Observe, that not every permutation of colors is valid.
For example, the tokens on positions 5, 14, 23, 32, 41, and 50 correspond to centers of faces, thus they are fixed.
There are more such constraints, but they are irrelevant to the pipeline itself.

\textbf{Action representation}
In our experiments we use quarter turns, i.e. an action corresponds to rotating a face by $90^{\circ}$, either clockwise or counterclockwise.
Since the action space contains only 12 elements, we use unique tokens to represent each of them.

\textbf{Low-level conditional policy input representation.}
The conditional policy takes two states $s$ and $s'$, which correspond to the current state and the state to be reached.
To represent such pairs, on every position we put a token corresponding to a pair of colors -- one located on that position in $s$ and the other in $s'$.
Since there are only 6 distinct colors on the Rubik's Cube, this requires using only 36 tokens.

\section{Training details}\label{sec:training_details_appendix}

\subsection{INT and Rubik's Cube}



\subsubsection{Transformer training}
For transformer training and inference we used HuggingFace's Transformers library \cite{DBLP:journals/corr/abs-1910-03771}. We did not use any pretrained checkpoints from HuggingFace model hub. We took mBART model class instead -- and trained it from scratch in a supervised way using HuggingFace's training pipeline. We generated (or loaded from disk) a fresh dataset for every epoch. Training batch was of size 32. For regularization, we set $dropout = 0.1$, but we did not use label smoothing (as opposed to \cite{vaswani2017attention}).

For the Adam optimizer we set $\beta_1 = 0.9$, $\beta_2 = 0.999$ and $\epsilon = 10^{-8}$. Learning rate schedule followed the formula:

$$lr = peak\_lr * \min\left(\frac{step\_num}{warmup\_steps}, \sqrt{\frac{warmup\_steps}{step\_num}}\right),$$
where $peak\_lr = 3 \cdot 10^{-4}$ and $warmup\_steps = 4000$.

The schedule curve matches the one proposed in \cite{vaswani2017attention}, but they use $peak\_lr \approx 7 \cdot 10^{-4}$.

\subsubsection{Sequence generation}

We used beam search with the number of beams set to 16 for INT and to 32 for Rubik's Cube. The number of returned sequences varied from 1 to 4 depending on the network type.

We set softmax temperature to $1$ by default. For Rubik's Cube subgoal generator we tuned this parameter and the value of $0.5$ performed best. We conducted a corresponding experiment for the baseline policy, but the temperature did not impact results in this case, as the policy outputs only a single token. For INT we did not tune the temperature, so we kept the value of $1$.

\subsubsection{Random seeds}\label{sec:seeds_appendix}


Due to use of the supervised learning, we observed little variance with respect to the random initialization. We tested this for the subgoal generator on proofs of length 10 and for $k=3$. Namely, we trained $5$ models of the subgoal generator, starting from different initializations. The success rate barely varied, as they stayed in the interval $[0.990, 0.992]$. In the other experiments, we used a single seed.

\subsection{Sokoban}

For training of the convolutional networks in Sokoban we set the learning rate to  $10^{-4}$ and the number of epochs to 200.

\newpage
\section{Wall-time for \abbrv{}}\label{sec:wall_time_appendix}
As indicated in Table \ref{table:int_success_rates}, \abbrv{} builds smaller search graphs. This has the practical advantage of making fewer neural network calls and consequently a substantially better wall-time.

The gains might be as high as $7$ times due to costly sequential calls of transformer networks, see Table \ref{table:int_wall_times}.




\begin{table}[h]{\small
\begin{tabular}{l|cc|cc|cc}
\toprule
Proof length           & \multicolumn{2}{|c|}{5} & \multicolumn{2}{c|}{10} & \multicolumn{2}{c}{15}                 \\ 
\midrule
Method             & 
{\scriptsize BestFS}   & {{\scriptsize  BF-kSubS {\tiny(ours)}}}   & {\tiny BestFS}     & {{\scriptsize BF-kSubS {\tiny(ours)}}}   & {\scriptsize BestFS}  & {{\scriptsize BF-kSubS {\scriptsize(ours)}}}  \\ 
\midrule
Total wall-time              & 4h 12m    & \textbf{3h 44m}     & 29h 22m    & \textbf{5h 55m}     & 69h 15m & \textbf{9h 22m}  \\ 
Avg. generator calls & NA & \textbf{3.04} &	NA & \textbf{3.89} &	NA & \textbf{6.23} \\
Avg. value calls & 23.34 & \textbf{4.01} & 112.35 & \textbf{4.80} & 159.46 & \textbf{6.59} \\
Avg. policy calls & 22.41 &	\textbf{8.43} & 112.21 &	\textbf{13.09} & 161.02 & \textbf{20.29} \\
\bottomrule
\end{tabular}}
\caption{\small Resources consumption for INT. We present evaluation on 1000 proofs and split into calls of subgoal generator network (used only in the subgoal search), value network and policy network (we report an average number of calls for a single proof).}
\label{table:int_wall_times}
\end{table}

\section{Training dataset size analysis}\label{section:dataset_sizes}

We tested how the success rate of BF-\abbrv{} on 12x12 Sokoban boards depends on the size of the training set. The full dataset consists of $125$k trajectories. We trained subgoal generator and value network on subsets consisting of $0.5$, $0.25$, $0.05$ and $0.01$ of all trajectories. The results are presented in Table \ref{table:dataset_sizes}.

\begin{table}[h]
\centering
\begin{tabular}{l|c|c|c|c|c}
\toprule
Fraction of the dataset           & 
 $1$    & $0.5$  & $0.25$   & $0.05$   & $0.01$  \\ 
\midrule
Success rate     & 0.93    & 0.86     & 0.84    &  0.48  & 0.14   \\ 
\bottomrule
\end{tabular}
\caption{\small Sokoban success rates for different training set sizes.}
\label{table:dataset_sizes}
\end{table}

\section{Value errors}\label{sec:appendix_value_noise}

\subsection{INT analysis}\label{sec:appendix_local_guidance}

Due to the size of state spaces, it is impossible to search over the entire space of INT formulas to find the shortest proofs. We instead analyze value estimations along proofs generated by INT engine. The monotonicity analysis in Section \ref{sec:local_guidance} was performed using $100$ such proofs of length $10$. The probabilities of value decrease for different step lengths $l$ are presented in Table \ref{table:value_decrease}.

\subsection{Sokoban Analysis}\label{sec:appendix_local_guidance}

\begin{wraptable}{R}{4cm}
\centering
\begin{tabular}{c|c}
\toprule
{\small $l$} & {\small Value decrease prob.} \\ \midrule
1           & 0.316               \\ 
2           & 0.217               \\ 
3           & 0.080               \\ 
4           & 0.020               \\ \bottomrule
\end{tabular}
\caption{}
\label{table:value_decrease}
\end{wraptable}

Here, we present details related to the Sokoban analysis from Section~\ref{sec:local_guidance}. We sampled $500$ Sokoban boards (with dimension $(12, 12)$). For each board, we calculated a full state-action graph and minimal distance from each state to the solution (this was needed to compute $S(s)$ sets later on). Since Sokoban graphs can be very large we set the limit on the graph size to 200000, which left us with 119 boards. Next, for each board, we took the sequence of states from the shortest solving path from the initial state (let us call this dataset as \textit{shortest-paths} - SP). For each pair of consecutive states in SP, we calculated the difference of the value estimation, and averaged them, which gave us a mean one-step improvement of $1.34$. We calculated this metric for 5 value function networks trained with different initialization, obtaining mean one-step improvement between $[1.23, 1.41]$.


To calculate the standard deviation of value function estimates for $S(s)$ we took SP, and limit it to states $s$ such that $|S(s)| \geq 5$ (lets denote it as SP5). We calculated standard deviation for each $s \in SP5$ separately. This gave us a mean deviation of $2.43$. (between $[2.24, 2.86]$ for 5 value networks trained with different initialization) The same set SP5 was used to calculate probabilities related to overoptimistic errors on Sokoban described at the end of Section \ref{sec:local_guidance}.

To calculate the above statistics we used the value function trained with supervised learning to approximate the distance to the solution. We observe that similar problems arise also when using value function trained with reinforcement learning \cite{milos2019uncertainty}. In such setup, mean variance of value function estimates for $S(s)$ is $0.84$, when one step improvement equals to $0.33$ . Probability that there is a state in $S(s)$ with value higher than best immediate neighbour of $s$ is $86$\% and it drops to $38$\%, if one considers states closer by $4$ steps.


\section{Example subgoals}\label{sec:example_subgoals_appendix}
\subsection{Example subgoals sets}
In this section, we present some example outcomes of the subgoal generator for INT and Sokoban. 

\subsubsection{INT}
In (\ref{example_int_1}) and (\ref{example_int_2}) we provide two examples of applying the subgoal generator (trained on proofs of length 5) to the given states in INT. The number of subgoals varies since not all of the outputs generated by the network could be reached by the conditional low-level policy.

\begin{align}\label{example_int_1}
\text{Input state: } &  {\scriptstyle (((b \cdot b)+((b+(b+f)) \cdot b))+(f+f))\geq ((((b+(b+b)) \cdot b)+0)+c)} \nonumber \\
\text{Ground truth 1: } &  {\scriptstyle (b+f)=b} \nonumber \\
\text{Ground truth 2: } &  {\scriptstyle(f+f)\geq c} \nonumber \\
\text{Subgoal 1: }& {\scriptstyle ((b \cdot b)+((b+(b+f)) \cdot b))=(((b+(b+b)) \cdot b)+0)}  \nonumber  \\
\text{Subgoal 2: }&  {\scriptstyle (((b+(b+f)) \cdot b)+(b \cdot b))=(((b+(b+b)) \cdot b)+0)} \nonumber  \\
\text{Subgoal 3: }&  {\scriptstyle ((b \cdot b)+((b+(f+b))  \cdot b))=(((b+(b+b)) \cdot b)+0)} \nonumber \\ 
\end{align}
\begin{align}\label{example_int_2}
\text{Input state: }  & {\scriptstyle ((((b \cdot (\frac{1}{b}))+a)^2)+(c+(\frac{1}{b})))=((((((\frac{1}{b}) \cdot b)+a) \cdot (a+1))+c)+(\frac{1}{b}))} \nonumber \\
\text{Subgoal 1: } & {\scriptstyle ((((b \cdot (\frac{1}{b}))+a)^2)+(c+(\frac{1}{b})))=((((b \cdot (\frac{1}{b}))+a) \cdot (a+1))+(c+(\frac{1}{b})))} \nonumber \\
\text{Subgoal 2: } & {\scriptstyle ((((b \cdot (\frac{1}{b}))+a) \cdot ((b \cdot (\frac{1}{b}))+a))+(c+(\frac{1}{b})))=((((b \cdot (\frac{1}{b}))+a) \cdot (a+1))+(c+(\frac{1}{b})))} \nonumber \\
\end{align}

\subsubsection{Sokoban}

Here we present two examples of the outcomes of the subgoal  generator trained for $12\times12$ boards:

\begin{figure}[h]
\centering
\includegraphics[height=0.22\textwidth]{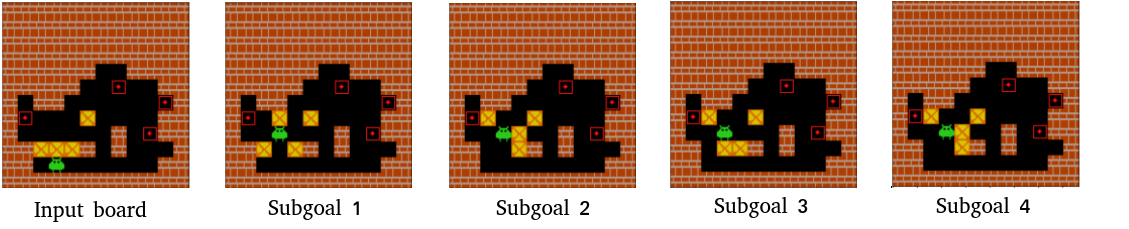}
\label{fig:sokoban_goals_1}
\end{figure}

\begin{figure}[h]
\centering
\includegraphics[height=0.22\textwidth]{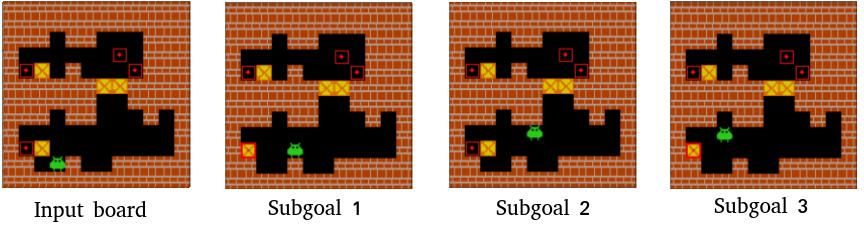}
\label{fig:sokoban_goals_2}
\end{figure}

\subsection{Example solutions with subgoals}

In this section, we present several examples of solutions obtained with our method. For simplicity, we only show the subgoal states on which the successful trajectories were constructed. In our setup, the last subgoal is always a solution.

\subsubsection{INT}

An example solution of INT problem of length 5:

$$\text{Problem: }{\scriptstyle (((0 \cdot ((a+0)+(-(a \cdot 1)))) \cdot (\frac{1}{(0^2)}))+((a+0)+(0^2))) \geq ((((0^2)+(1+(a+0)))+b)+(-((a+0)+f)))} $$
\begin{align}
\text{1st subgoal: } & {\scriptstyle (((0 \cdot ((a+0)+(-(a \cdot 1)))) \cdot (\frac{1}{(0^2)}))+((a+0)+(0^2)))=((0^2)+(1+(a+0)))}  \nonumber \\
\text{2nd subgoal: } & {\scriptstyle (((0 \cdot ((0+a)+(-(a \cdot 1)))) \cdot (\frac{1}{(0^2)}))+((a+0)+(0^2)))=(1+((a+0)+(0^2)))}  \nonumber \\
\text{3rd subgoal: } & {\scriptstyle (0+a)=(a \cdot 1)}  \nonumber \\
\text{4th subgoal: } & {\scriptstyle a=a}  \nonumber \\
\end{align}

\subsubsection{Sokoban}

An example solution of Sokoban board:

\begin{figure}[h]
\centering
\includegraphics[width=0.90\textwidth]{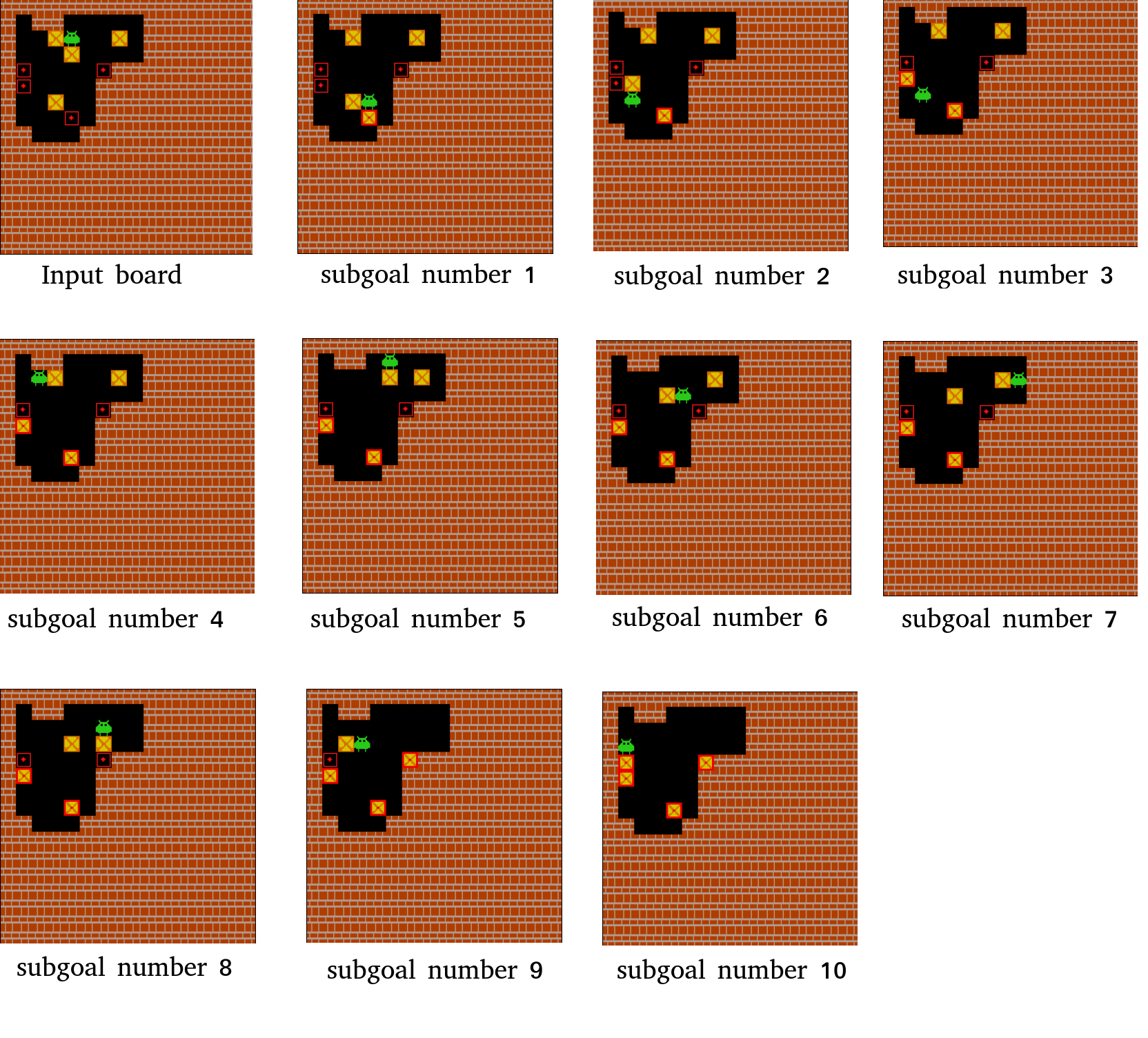}
\label{fig:sokoban_solution}
\vspace{-0.5cm}
\end{figure}

\section{Baselines} \label{sec:baslines_app}

\begin{wrapfigure}{R}{0.51\textwidth}
\begin{minipage}[h]{.51\textwidth}
\begin{algorithm}[H]\label{alg:low_level_generator}
    \caption{Low-level generator}
    \label{alg:low_level_generator}
\begin{tabular}{ l c l }
    \textbf{Require: }
    & $C_3$ & number of states to produce \\
    & $NB$ & number of beams in sampling\\
    & $\pi_b$ & behavioral cloning policy \\
    & $M$ & model of the environment \\
\end{tabular}
\begin{algorithmic}
    \Function{sub\_generate}{$\mathtt{s}$}
        \State $\mathtt{actions} \gets \Call{beam\_search}{\pi_b,\mathtt{s}; C_3, NB}$
        \State $\mathtt{subgoals} \gets []$
        \For{$\mathtt{action} \in \mathtt{actions}$}
            \State $s \gets M.\Call{next\_state}{\mathtt{s}, \mathtt{action}}$
            \State $\mathtt{subgoals}.\Call{append}{\mathtt{s}}$
        \EndFor
        \State \Return $\mathtt{subgoals}$
    \EndFunction
\end{algorithmic}
\end{algorithm}
\end{minipage}
\end{wrapfigure}

Our first baseline is the low-level policy trained with behavioral cloning from the expert data trained to solve the problem (contrary to the low-level conditional policy $P$, which aims to achieve subgoals). Such policy was used \cite{wu2020int}. We verified that our behavioral cloning policy reproduces the results from \cite{wu2020int} for proofs of lengths $5$. 

\textbf{MCTS}. As a baseline for MCT-\abbrv{} we used an AlphaZero-based MCTS planner described in Appendix \ref{sec:mcts_algorithm_appendix}. 

\textbf{BestFS}. The baseline for BF-\abbrv{} is a low-level planning. We substitute $\Call{sub\_generate}{}$ with a function returning adjacent states indicated by the most probable actions of behavioral cloning policy, see Algorithm \ref{alg:low_level_generator}. 

\section{Simple planners}\label{sec:appendix_simple_planners}

An appropriate search mechanism is an important design element of our method. To show this, we evaluate an alternative, simpler procedure used by e.g. \cite{fang2019dynamics} for subgoal-based planning. It works by sampling independent sequences of subgoals and selects the best one. This method solved none of 1000 Rubik's Cube instances despite using the same subgoal-generator as BF-\abbrv{} (which has a success rate of $0.999$ with a comparable computational budget).

\section{Investigation of baseline-BestFS on Rubik's Cube}\label{sec:appendix_rubik_baseline}

To obtain the training datasets on the Rubik's Cube environment, we generated random paths starting from a solved cube and stored them in the reversed order. These backward solutions are highly sub-optimal: for example, the states obtained by 10 random moves are usually in a distance of about 6 - 7 steps from the final state and the gap gets much larger for a higher number of moves. This means that on collected trajectories only some of the actions indeed lead to the solution and the majority of them only introduce noise.

We observed that the baseline is much weaker than BF-\abbrv{} even for $k=1$, despite the effort on tuning it. We extended the training of behavioral cloning policy and did a  grid-search over parameters to further improve its success rate. We managed to reach no more than 10\% success rate for the best version. To provide a meaningful evaluation of the baseline, we also trained it on very short trajectories consisting of 10 random moves. Such a curriculum allowed the behavioral cloning policy to learn, however still suffered from randomness in data (for states located far from the solution).

Interestingly, BF-\abbrv{} for $k=1$ turned out to perform much better than the baseline. Full understanding of this phenomenon requires additional research, however we hypothesize that in our setup learning to predict states is an easier task that predicting an action. A potential reasons are that: states prediction provides a denser learning signal and Transformers perform better when dealing with sequences (then with predicting a single token).

Note that both \abbrv{} and baseline policy learn from fully off-policy data. The problem of solving the Rubik's Cube is challenging, thus learning from noisy off-policy trajectories can be simply too hard for standard algorithms.


\section{Technical details}\label{sec:techical_details}
\subsection{Infrastructure used} \label{sec:infrastructure_used}

We had 2 types of computational nodes at our disposal, depending on whether a job required GPU or not. GPU tasks used a single Nvidia V100 $32$GB card (mostly for transformer training) and Nvidia RTX 2080Ti $11$GB (for evaluation) with 4 CPU cores and 16GB RAM. The typical configuration of CPU job was the Intel Xeon E5-2697 $2.60$GHz processor (28 cores) with $128$GB memory.

Each transformer for INT was trained on a single GPU for 3 days (irrespective of proof length and the network type). INT evaluation experiments used a single GPU for a time period varying from several hours to 3 days -- with baselines being the bottleneck.

Transformers for Rubik's Cube required more training -- every network was trained for 6 days on a single GPU. Because of relatively short sequences representing the cube's state, we were able to run evaluations without GPU. We used 20 CPU cores with $20$GB memory, while still fitting in a 3-day run time.

We trained and evaluated Sokoban on CPU only
mostly because of rather small neural network sizes. Training time varied from 2 to 3 days, whereas evaluation took only an hour.


\end{document}